\documentclass[10pt,twocolumn,letterpaper]{article}

\usepackage{misk/cvpr}              

\usepackage{graphicx}
\usepackage{amsmath}
\usepackage{bm}
\usepackage{makecell}
\usepackage{tabularx}
\usepackage{multirow}
\usepackage{amssymb}
\usepackage{amsmath}
\usepackage{tabularx}
\usepackage{bm}
\usepackage[ruled,linesnumbered]{algorithm2e}
\usepackage{algorithmic}
\usepackage{xcolor}
\usepackage{listings}
\newcolumntype{Y}{>{\centering\arraybackslash}X}
\usepackage{arydshln}
\usepackage{multirow}
\usepackage{pifont}

\newcommand{\cmark}{\ding{51}}
\newcommand{\xmark}{\ding{55}}

\renewcommand{\paragraph}[1]{\vspace{.5em}\noindent\textbf{#1.}}

\definecolor{cvprblue}{rgb}{0.21,0.49,0.74}
\usepackage[pagebackref,breaklinks,colorlinks,allcolors=cvprblue]{hyperref}

\def\paperID{*****} 
\def\confName{CVPR}
\def\confYear{2026}

\newcommand{\tocite}[1]{\textcolor{blue}{[TO CITE]}}
\newcommand{\ourmethod}{{DiverseDiT}\xspace}

\title{DiverseDiT: Towards Diverse Representation Learning in Diffusion Transformers}

\author{
Mengping Yang$^{1,2}$ \enspace
Zhiyu Tan$^{1,2\dagger}$ \enspace
Binglei Li$^{1,2,3}$ \enspace
Xiaomeng Yang$^{2}$ \enspace
Hesen Chen$^{1,2}$ \enspace
Hao Li$^{1,2,3*}$ \\[2pt]
\textsuperscript{1}Fudan University \quad
\textsuperscript{2}Shanghai Academy of AI for Science \quad
\textsuperscript{3}Shanghai Innovation Institute \\
}
\begin{document}

\twocolumn[{
\maketitle
\begin{center}
    \captionsetup{type=figure}
    \includegraphics[width=.95\linewidth]{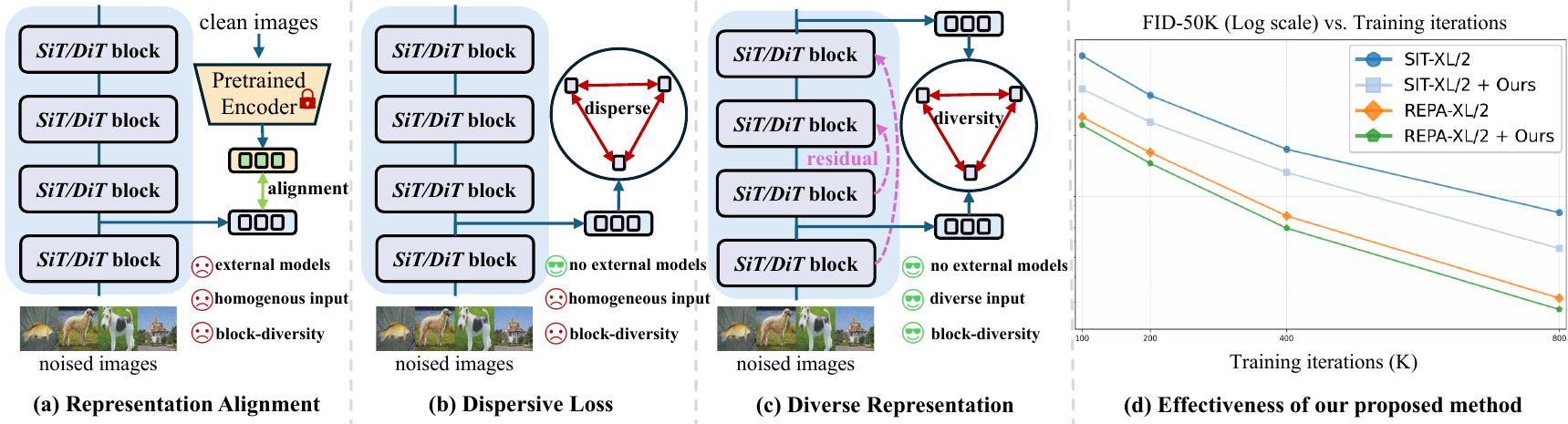}
    \vskip -0.1in
    \caption{
    \textbf{Comparison between Representation Alignment~\cite{repa}, DispLoss~\cite{wang2025diffuse} and our proposed DiverseDiT in learning representations}.
    (a) REPA~\cite{repa} employs external encoders as guidance and different blocks'inputs are homogeneous.
    (b) DispLoss~\cite{wang2025diffuse} encourage internal representations to spread out but still with homogeneous input and without block-wise diversity.
    (c) We propose long residual connections to enhance input diversity and diversity loss to encourage diverse feature representations across blocks.
    (d) On ImageNet 256 $\times$ 256, our proposed method consistently reflects training efficiency and effectiveness when applied to both SiT and REPA.
    }
     \label{fig:teaser}
\end{center}
}
]

\maketitle

{
  \renewcommand{\thefootnote}%
    {\fnsymbol{footnote}}
  \footnotetext{\textsuperscript{$\dagger$}Project Lead, \textsuperscript{*}Corresponding Author}
}

\begin{abstract}
Recent breakthroughs in Diffusion Transformers (DiTs) have revolutionized the field of visual synthesis due to their superior scalability.
To facilitate DiTs' capability of capturing meaningful internal representations, recent works such as REPA incorporate external pretrained encoders for representation alignment.
However, the underlying mechanisms governing representation learning within DiTs are not well understood.
To this end, we first systematically investigate the representation dynamics of DiTs. Through analyzing the evolution and influence of internal representations under various settings, we reveal that representation diversity across blocks is a crucial factor for effective learning.
Based on this key insight, we propose \ourmethod, a novel framework that explicitly promotes representation diversity. 
\ourmethod incorporates long residual connections to diversify input representations across blocks and a representation diversity loss to encourage blocks to learn distinct features.
Extensive experiments on ImageNet 256 $\times$ 256 and 512 $\times$ 512 demonstrate that our \ourmethod yields consistent performance gains and convergence acceleration when applied to different backbones with various sizes, even when tested on the challenging one-step generation setting.
Furthermore, we show that \ourmethod is complementary to existing representation learning techniques, leading to further performance gains. 
Our work provides valuable insights into the representation learning dynamics of DiTs and offers a practical approach for enhancing their performance.
Our code is available at \url{https://github.com/kobeshegu/DiverseDiT}.
%
\end{abstract}    
\section{Introduction}
\label{sec:intro}

%
Diffusion models~\cite{ddpm,sohl2015deep}, particularly diffusion transformers (DiT)~\cite{dit,bao2023all} which demonstrate superior scalability in learning the data distribution, have significantly advanced the field of visual synthesis including text-to-image~\cite{xie2025sana,rombach2022high}, text-to-video generation~\cite{yang2024cogvideox,gupta2024photorealistic}, \emph{etc}.

Recent studies identified that top-performing diffusion models capture more discriminative internal representations~\cite{mittal2023diffusion, chen2024deconstructing, xiang2023denoising}, yielding an implicit connection between diffusion generative models and representation learning.
Following this philosophy, methods like REPA~\cite{repa} (\cref{fig:teaser} (a)) align latent noisy representations with features derived from pre-trained visual encoders to guide the representation learning. 
Subsequent work REPA-E~\cite{leng2025repa} extends such alignment in a joint end-to-end training manner with VAE tuning and REG~\cite{wu2025representation} entangles low-level visual latents and high-level class tokens for a two-level alignment.
However, this reliance on powerful external foundation models, which require massive resources for training, poses a significant drawback.
Other approaches aim to improve representations without external guidance. 
SRA~\cite{jiang2025no} aligns representations between a student and an EMA teacher model, while Wang \etal~\cite{wang2025diffuse} propose a dispersive loss to encourage separation between internal representations (\cref{fig:teaser} (b)).
Despite their considerable advancements, the underlying mechanisms governing representation learning within DiT models remain largely opaque.
Key questions persist: 
How do DiT models learn meaningful representations, and why are external alignment techniques effective? 
This lack of fundamental understanding hinders the development of more principled and efficient training paradigms.

To address this gap, we perform a systematic investigation into the representation learning process of DiT models (\cref{sec:revisting_repa}).
First, we analyze \textit{how representations evolve throughout training} by measuring the discrepancy between internal representations across different blocks.
Then, we examine \textit{how representations are influenced} when external models are employed for alignment on \textit{different blocks} with \textit{multiple encoders}.
Through these analyses, we reveal several key findings:
1) Representation discrepancy across blocks naturally increases as training progresses;
2) Aligning a single block with a pre-trained model significantly increases its discrepancy from other blocks;
3) Crucially, aligning more blocks or using more encoders does not necessarily improve performance, suggesting that excessive alignment can harm the overall diversity of the model's representations. 
These observations provide a new perspective for understanding the representation learning of DiTs and offer a plausible explanation for the effectiveness of existing techniques like REPA: the rationale behind effective representation learning in DiTs lies in \textit{improving the representation diversity} across different blocks.

Capitalizing on the above insights, we propose \ourmethod, a novel and effective framework for promoting diverse representation learning in DiTs.
Specifically, \ourmethod introduces two simple yet powerful components. 
First, we incorporate long-range residual connections to diversify the inputs to different blocks, preventing representational homogenization. 
Second, we introduce a representation diversity loss that explicitly penalizes similarity between features from different blocks. This encourages each block to specialize and capture unique, complementary aspects of the data. 
Together, these components promote diverse representation learning through both diverse inputs and inter-block diversity constraints, without requiring external guidance models (\cref{fig:teaser} (c)).
We conduct extensive experiments to testify the effectiveness of \ourmethod on ImageNet 256$\times$256 and ImageNet 512$\times$512 datasets. 
The results demonstrate that our method consistently improves the training convergence and synthesis quality when applied to different baselines, with or without external guidance (\cref{fig:teaser} (d)) across a wide range of model scales, even when tested on the challenging one-step setting~\cite{geng2025mean}.
Moreover, we show that our proposed method is complementary to existing alternatives such as Disp~\cite{wang2025diffuse} and SRA~\cite{jiang2025no}, yielding further improvement.

We summarize our primary contributions as:
1) We conduct a comprehensive analysis on how representations are learned in DiTs, revealing that improving representation diversity across blocks is a key factor for effective training. 
To our knowledge, this work is the \textit{first to elucidate this representation relationship} and provide valuable insight to understanding representation dynamics of DiTs.
2) We propose \ourmethod, an efficient and effective framework that facilitates representation diversity with long residual connections that enable diverse input and block diversity loss that encourages representations to be distinct.
3) Extensive experiments with different baseline models across various scales on both multi-step and one-step settings demonstrate the effectiveness of our method in accelerating convergence and improving performance.

\section{How Representations Are Learned?}
\label{sec:revisting_repa}

\begin{figure*}[t]
    \vskip -0.2in
    \begin{center}  \centerline{\includegraphics[width=.95\linewidth]{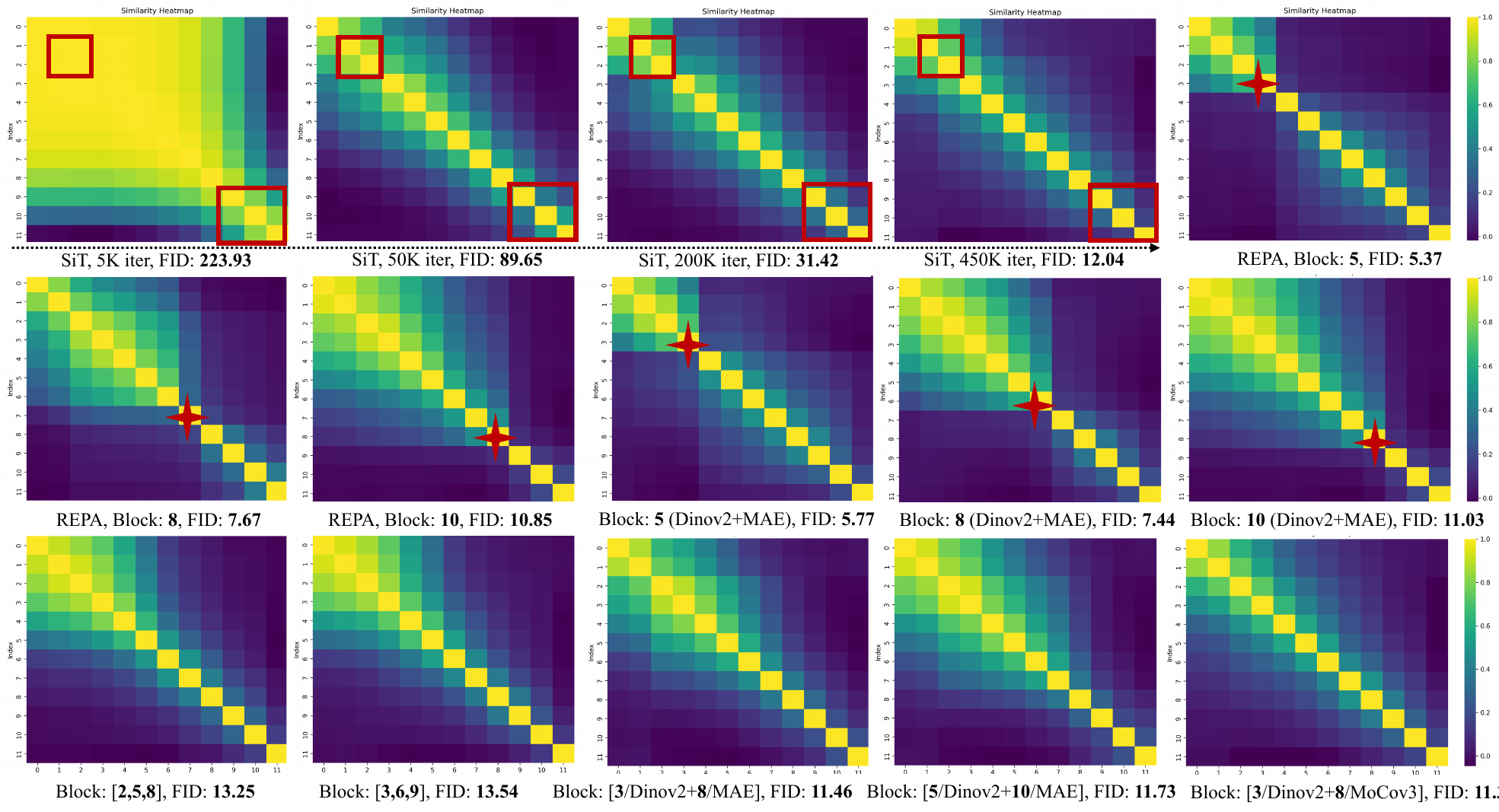}}
    \vskip -0.1in
    \caption{\textbf{CKA representation similarities of models trained on various settings.}  
    We can observe that 1) the discrepancies between different blocks increases as training progresses;
    2) aligning specific blocks significantly increases the dissimilarity between the corresponding block and other blocks;
    3) aligning on more blocks with different pretrained encoders brings marginal performance improvements.
    Detailed quantitative results are provided in \cref{sec:supp:analysis_details}.
    }
    \label{fig:analysis}
    \end{center}
    \vskip -0.4in
\end{figure*}

\subsection{Preliminaries}
\label{sec:preliminaries}

\noindent \textbf{Scalable Interpolant Transformers (SiT).}
Our work is based on SiT, which unifies flow~\cite{flowmatching_lipman2022} and diffusion~\cite{ddpm} models that transform Gaussian noise $\epsilon$ into samples $\mathbf{x}_*$:
\begin{equation}
    \mathbf{x}_t = \alpha_t \mathbf{x}_* + \sigma_t \epsilon,
\end{equation}
where $\alpha_t$ decreases and $\sigma_t$ increases with time $t$.
Flow-based models interpolate between noise and data while diffusion models define a stochastic differential equation to approach Gaussian distribution as $t\rightarrow \infty$.
Sampling is performed via a reverse SDE for diffusion or a probability flow ODE for flow models: $\dot{\mathbf{x}}_t = \mathbf{v}(\mathbf{x}_t, t)$, the velocity field $\mathbf{v}(\mathbf{x}_t, t)$ can be formulated with conditional expectation:
\begin{equation}
    \mathbf{v}\!(\mathbf{x}\!,\!t)\! \!=\! \mathbb{E}[\dot{\mathbf{x}}_t \!\mid\! \!\mathbf{x}_t\! \!=\! \mathbf{x}] \!\!=\!\! \dot{\alpha}_t \mathbb{E}[\mathbf{x}_* \!\mid\! \!\mathbf{x}_t\! \!=\! \!\mathbf{x}] \!\!+\! \dot{\sigma}_t \mathbb{E}[\epsilon \!\!\mid\!\! \mathbf{x}_t\! \!=\! \!\mathbf{x}].
\end{equation}
The velocity for the velocity field $\mathbf{v}(\mathbf{x}_t, t)$ is derived by a model $\mathbf{v_\theta}(\mathbf{x}_t, t)$ trained to minimize:
\begin{equation}
    \mathbb{E}_{\mathbf{x}_*, \epsilon, t} \left[ \| \mathbf{v}_\theta(\mathbf{x}_t, t) - \dot{\alpha}_t \mathbf{x}_* - \dot{\sigma}_t \epsilon \|^2 \right].
\end{equation}
Once trained, we can synthesize samples from random noises by the reverse SDE via computing the velocity filed:
\begin{equation}
d\mathbf{x}_t \!=\! \mathbf{v}(\mathbf{x}_t, t) dt \!-\! \frac{1}{2} w_t \mathbf{s}(\mathbf{x}_t, t) dt + \sqrt{w_t} d\overline{\mathbf{w}}_t,
\end{equation}
where score $\mathbf{s}(\mathbf{x}_t, t)$ is obtained via conditional expectation:
\begin{equation}
\mathbf{s}(\mathbf{x}_t, t) \!=\! -{\sigma}_t^{-1} \mathbb{E}[\epsilon \!\mid\! \mathbf{x}_t \!=\! \mathbf{x}] \!=\! \sigma_t^{-1} \frac{\alpha_t \mathbf{v}(\mathbf{x}, t) - \dot{\alpha}_t \mathbf{x}}{\alpha_t \dot{\sigma}_t - \dot{\alpha}_t \sigma_t}.
\end{equation}

\noindent \textbf{Representation Alignment (REPA).}
To leverage external models to aid representation learning for DiTs, REPA proposes to perform patch-wise projection alignment between the model's intermediate hidden states $h$ with features $\mathbf{y}_*$ derived from pretrained visual encoders:
\begin{equation}
\mathcal{L}_{\text{REPA}}(\!\theta,\! \phi)\! \!:=\! -\mathbb{E}_{\mathbf{x}_*, \epsilon, t} \!\left[\! \frac{1}{N} \sum_{n=1}^{N} \text{sim}(\mathbf{y}_*^{[n]}, h_\phi(\mathbf{h}_t^{[n]})) \!\right],
\end{equation}
where $\mathbf{x}_*$ denotes clean images, $h_\phi$ is MLP projectors and $\text{sim}(\cdot,\cdot)$ denotes similarity function.

\noindent \textbf{Centered Kernel Alignment (CKA).}
CKA is a widely used similarity index for quantifying neural network representations~\cite{cristianini2001kernel, kornblith2019similarity, davari2022reliability}.
Accordingly, we adopt CKA to calculate the similarities of representations across DiT blocks for our analysis.
Formally, CKA is normalized from Hilbert-Schmidt Independence Criterion (HSIC)~\cite{HSIC} to be invariant to orthogonal
transformation and isotropic scaling:
\begin{align}
    \label{eq:cka}
    \mathrm{CKA(X,Y)}=\frac{\mathrm{HSIC}(\mathrm{x},\mathrm{y})}{\sqrt{\mathrm{HSIC}(\mathrm{x},\mathrm{x}) \mathrm{HSIC}(\mathrm{y},\mathrm{y})}}.
\end{align}
HSIC identifies whether two distributions ($\mathrm{X,Y}$) are independent: $\mathrm{HSIC}(K,L) \!=\! \frac{1}{(n-1)^2}\operatorname{Tr}(K H L H)$, $K_{i j}\!=\!k\left(\mathrm{x}_i, \mathrm{x}_j\right)$ { and } $L_{i j}\!=\!l\left(\mathrm{y}_i, \mathrm{y}_j\right)$, where $k$ and $l$ are kernels.

\subsection{Our Observations}
\label{sec:observations}
With CKA as the representation similarity index, we systematically investigate how representations are learned and how they are affected when external representation alignment is enabled in three settings.
%
1) SiT training stage analysis: we track the evolution of internal representations and quantify the change in similarity between different blocks as the model learns.
2) REPA block-specific alignment: we identify the effect of aligning pretrained visual features on different blocks to assess how external knowledge alters the representation of specific blocks.
3) REPA multiple block guidance from multiple encoders: we explore the impact of applying external guidance to multiple blocks with multiple encoders to probe whether guiding multiple blocks leads to improvement.
All implementation details strictly follow the settings of SiT-B/2 and REPA-B/2 on Imagenet $256\times256$ for 450K iterations.
We use DINOv2-B, MAE-L and MoCov3 for REPA alignment.
The visualized results are shown in \cref{{fig:analysis}} and detailed quantitative results are given in \cref{sec:supp:analysis_details}, we can observe several interesting findings from the results.
%

\noindent \textbf{(1). Representation diversity across different blocks increases during training:}
The similarity heatmaps of SiT at different training steps (5K, 50K, 200K, 450K) show a clear trend of increasing representational diversity.
Specifically, the heatmap becomes more diagonal as training progresses, and the representation between different layers becomes less similar.
Intuitively, different blocks specialize, develop more distinct and complementary representations. 
Such observation aligns with the broader understanding that deep models learn hierarchical representations.
%

\noindent \textbf{(2). External alignment enhances block differentiation:}
The REPA heatmaps exhibit more distinct (less similar) patterns around the red mark compared to the corresponding regions in the SiT heatmaps, indicating that aligning specific blocks significantly increases the dissimilarity between the representations of the targeted block and other blocks.
Additionally, consistent with REPA~\cite{repa}, aligning earlier blocks (\emph{i.e.,} Block 5, Block 8) yields better performance than aligning later blocks (Block 10). 
This demonstrates that external alignment effectively promotes specialization by making the selected block's representation more different from other blocks. 
In other words, REPA encourages each block to learn more distinct and complementary features, leading to a more diverse and more effective representation. 
More importantly, these observations provide insight into why REPA-like external alignment is effective: by enforcing specialization, it prevents representational collapse and encourages the network to explore a wider range of features.
This specialization-driven perspective may also explain why aligning with larger models (DINOv2-L, -g) brings only marginal improvements compared to aligning with smaller models (DINOv2-B) in the original REPA.
%

\noindent \textbf{(3). Aligning on more blocks with more external models does not necessarily improve performance:}
While REPA with single blocks shows clear differentiation, using multiple blocks for guidance (e.g., Block:[2,5,8], [3,6,9]) does not bring similar improvements to the performance. 
In some cases, the FID score is even slightly worse (Block [2,5,8]), suggesting that applying guidance to more blocks might counterintuitively reduce the overall diversity between blocks.
We hypothesize that this is due to the introduction of conflicting constraints, preventing individual blocks from effectively specializing. 
Furthermore, aligning multiple blocks with different external encoders ([5/Dinov2+10/MAE], \emph{i.e.}, aligning DinoV2 features on block 5 and MAE features on block 10) also provides limited benefit and shows limited representation diversity.
Such observation further reflects that the representation diversity across blocks is a crucial factor for high-quality synthesis.
%

In general, our systematic analysis provides a comprehensive understanding of representation dynamics for DiTs and reveals that the key for representation learning is increasing the discrepancies of block representations.
Our findings offer a novel perspective for explaining existing methods, showing how models learn representations during training and highlighting the critical role of block specialization.
These observations motivate us to design more effective methods to enhance representation diversity for performance improvement and accelerated training.

\section{Methodology}
\label{sec:method}

In light of the observations from our systematic analysis of representation dynamics in \cref{sec:revisting_repa}, we introduce, \ourmethod, a novel method to explicitly enhance representation diversity.
Our approach focuses on encouraging specialization via long residual connections to enhance the input diversity of different blocks and a representation diversity loss to explicitly promote diverse feature representations across all blocks.
We detail each of them below.

\subsection{Long Residual Connections}
Motivated by the findings in \cref{sec:revisting_repa}, we argue that the diversity of inputs for each block also plays a crucial role in shaping the learned representations.
However, conventional diffusion transformers often suffer from a lack of input diversity because each block's input is typically homogeneous and is derived solely from the output of the preceding layer.
To address this, we employ a long residual connection to inject diversity into the inputs of each block.  
This mechanism selectively injects the output of earlier layers into later layers, promoting feature reuse and preventing representational collapse.
Formally, suppose the model consists of $L$ DiT blocks, we connect the $i$-th block's output to the ($L-i$)-th block via:
\begin{equation}
    {f}_l \!=\! \mathcal{R}_{\text{res}}^i({f}_i, {f}_{l-1}) \!=\! \text{Linear}(\text{Norm}({f}_i \oplus {f}_{l-1})),
\end{equation}
where $i \in [{0,...,L//2-1}]$, and $\mathcal{R}_{\text{res}}^i$ denotes the residual connection.
${f}_i \in \mathbb{R}^{N \times T \times D}$ is the representation of the $i$-th block, $\oplus$ denotes concatenating the representation of two blocks ${f}_i$ and ${f}_{l-1}$, which is further processed by a layer normalization and a linear layer for linear transformation.
By injecting skip connections, we break the chain of homogeneous inputs and encourage the network to learn more varied and informative representations from different sources.

\subsection{Representation Diversity Loss}
To further encourage specialization and promote diversity in the learned representations, we introduce a representation diversity loss to explicitly promote diverse feature representations within each block. 
Specifically, our representation diversity comprises three key components:
an orthogonality loss, a mutual-information minimization loss, and a feature dispersion loss.
%
%
Notably, to reduce computational cost, we only consider a subset of all possible pairs from $L$ blocks: $\mathcal{P} \subseteq \{(i, j) : i < j,\; i, j \in {L}\}$.
For each block, we define token-wise mean feature along the $N$ and $T$ dimension as:
\begin{equation}
    \bm{\mu}_l = \frac{1}{NT} \sum_{n=1}^{N} \sum_{t=1}^{T} f_l[n, t, :] \in \mathbb{R}^{D}.
\end{equation}
Then, we compute the \textit{orthogonality loss} by penalizing high cosine similarity between block-wise mean representations, encouraging cross-block orthogonality:
\begin{equation}
    \mathcal{L}_{\mathrm{orth}}  \!=\! \frac{1}{|\mathcal{P}|}\! \!\sum_{(i, j)\! \!\in \mathcal{P}}\! cos(\bm{\mu}_i, \bm{\mu}_j) \!=\! \frac{1}{|\mathcal{P}|} \!\sum_{(i, j)\! \in \mathcal{P}} \frac{\bm{\mu}_i^{\top} \, \bm{\mu}_j}{\lVert \bm{\mu}_i \rVert_2 \lVert \bm{\mu}_j \rVert_2}.
\end{equation}

Next, we minimize mutual information between block representations to ensure statistical independence within block-wise representations.
However, directly computing mutual information is computationally intractable for high-dimensional features.
Therefore, we use a computationally efficient proxy based on the average cosine similarity of normalized feature vectors as the estimation of mutual information.
Specifically, we define flattened, $\ell_2$-normalized token representations along the $N$ and $T$ dimension as:
\begin{equation}
    \hat{\bm{f}}_{l, n, t} = \frac{f_l[n, t, :]}{\lVert f_l[n, t, :] \rVert_2} \in \mathbb{R}^{D} .
\end{equation}
Then, we compute the \textit{proxy mutual-information loss} as:
\begin{equation}
    \mathcal{L}_{\mathrm{MI}} = \frac{1}{|\mathcal{P}|} \sum_{(i, j) \in \mathcal{P}} \frac{1}{NT} \sum_{n=1}^{N} \sum_{t=1}^{T} \hat{\bm{f}}_{i, n, t}^{\top} \hat{\bm{f}}_{j, n, t} .
\end{equation}
In this way, we avoid directly calculating covariance matrices for efficiency and meanwhile minimizing the correlation between representations.
Further, we employ a feature dispersion loss to encourage diverse channel usage by maximizing the variance of feature activations.
The representations of each block are flattened to $\tilde{f}_l \in \mathbb{R}^{(NT) \times D}$ and normalized along the sample axis to obtain $\widehat{\tilde{f}}_b$.
Then we compute the averaged activation per dimension:
\begin{equation}
    {a} = \frac{1}{|\mathcal{P}|} \sum_{p \in \mathcal{P}} \operatorname{mean}_{n, t}( \widehat{\tilde{f}}_b[n, t, :]),
\end{equation}
${a}$ is then normalized to ${a}'$ by ${a}' = {a} / \max_k a_k$, and its variance is maximized to obtain the \textit{feature dispersion loss}:
\begin{equation}
    \mathcal{L}_{\mathrm{disp}} = -\frac{1}{D} \sum_{k=1}^{D} (a'_k - \bar{a}')^2, \bar{a}' = \frac{1}{D} \sum_{k=1}^{D} a'_k .
\end{equation}
Finally, the \textit{overall representation diversity loss} aggregates the above three components as:
\begin{equation}
    \mathcal{L}_{\text{div}} = \lambda_{\text{orth}}\mathcal{L}_{\text{orth}} + \lambda_{\text{MI}}\mathcal{L}_{\text{MI}} + \lambda_{\text{disp}}\mathcal{L}_{\text{disp}},
\end{equation}
where $\lambda_{\text{orth}}, \lambda_{\text{MI}}, \lambda_{\text{disp}}$ control the relative weight of each loss, we set them as $0.33$ in default without any parameter searching.
In practice, we find that when $\mathcal{L}_{\text{div}}$ is optimized too small (\emph{e.g.,} close to 0), the model tends to diverge and becomes unable to effectively model the underlying data distribution.
This phenomenon potentially arises because overly emphasizing the separation of representations hinders the model's ability to specialize and learn meaningful, shared representations across the data.
We thus develop an adaptive weight $w$ for the overall $\mathcal{L}_{\text{div}}$:
$$
w =
\begin{cases}
  1, & \text{if } \mathcal{L}_{\text{div}} > 0.5, \\
  \frac{\mathcal{L}_{\text{div}} - 0.1}{0.5}, & \text{if } 0.1 < \mathcal{L}_{\text{div}} \leq 0.5, \\
  0, & \text{otherwise}.
\end{cases}
$$
%

\section{Experiments}
\label{sec:exps}

\begin{table}[t]
    \centering
    \small
    \vskip -0.1in
    \caption{\textbf{Variation in model-scale on ImageNet 256$\times$256 without CFG}. 
    Our proposed method brings consistent performance gains across all model-scales when applied to both SiT and REPA. 
    %
    }
    \vskip -0.1in
    \setlength{\tabcolsep}{1.8mm}{
    \begin{tabular}{lcccccc}
        \toprule
        \textbf{Model}  & \textbf{Iter}. & \textbf{FID}$_\downarrow$ & \textbf{sFID}$_\downarrow$ & \textbf{IS}$_\uparrow$ & \textbf{Prec.}$_\uparrow$ & \textbf{Rec.}$_\uparrow$ \\
        \midrule
        SiT-B             & 400k  & 36.80          & 6.77          & 40.09          & 0.51          & \textbf{0.63} \\
        \textbf{+ (Ours)} & 400k  & \textbf{28.05} & \textbf{6.04} & \textbf{50.66} & \textbf{0.57} & \textbf{0.63} \\
        \hdashline
        REPA-B            & 400k  & 22.99          & 6.70          & 64.73          & 0.59          & \textbf{0.65} \\
        \textbf{+ (Ours)} & 400k  & \textbf{17.29} & \textbf{6.56} & \textbf{79.92} & \textbf{0.62} & \textbf{0.65} \\
        \midrule
        SiT-L             & 400k  & 18.77          & 5.27          & 71.44          & 0.64          & 0.63 \\
        \textbf{+ (Ours)} & 400k  & \textbf{16.10} & \textbf{5.08} & \textbf{79.47} & \textbf{0.66} & \textbf{0.64} \\
        \hdashline
        REPA-L            & 400k  & 9.57           & \textbf{5.34} & 113.32         & \textbf{0.69} & 0.66          \\
        \textbf{+ (Ours)} & 400k  & \textbf{8.47}  & 5.42          & \textbf{123.03}& \textbf{0.69} & \textbf{0.67} \\
        \midrule
        SiT-XL            & 400k  & 17.43          & 5.11          & 76.00          & 0.64          & \textbf{0.64} \\
        \textbf{+ (Ours)} & 400k  & \textbf{12.42} & \textbf{4.85} & \textbf{95.01} & \textbf{0.68} & 0.63 \\
        \hdashline
        REPA-XL           & 400k  & 8.73           & 5.21          & 118.68         & 0.69          & \textbf{0.65} \\
        \textbf{+ (Ours)} & 400k  & \textbf{8.09}  & \textbf{5.02} & \textbf{123.23}& \textbf{0.70} & \textbf{0.65} \\
        \bottomrule
    \end{tabular}
    }
    \vskip -0.2in
    \label{tab:different_size}
\end{table}

\subsection{Experiment Setup}

\noindent \textbf{Implementation Details}
We incorporate \ourmethod into several popular baselines to evaluation its versatility and effectiveness including SiT~\cite{sit}, REPA~\cite{repa} and MeanFlow~\cite{geng2025mean}, leaving other details untouched.
To ensure fair comparison, we strictly follow the training configurations of SiT~\cite{sit}, REPA~\cite{repa} and MeanFlow~\cite{geng2025mean} for experimental evaluation on the ImageNet~\cite{deng2009imagenet} dataset with the resolution of 256$\times$256 and 512$\times$512, images and pre-computed VAE features are preprocessed following REPA~\cite{repa} with Stable Diffusion VAE~\cite{rombach2022high}.
We adapt the B/2, L/2, and XL/2 model configurations from SiT with a patch size of 2 to evaluate the scalability.
For training, we use AdamW~\cite{loshchilov2017decoupled} with a constant learning rate of 1e-4, ($\beta_1$, $\beta_1$) = (0.9, 0.999) without decay, the batchsize is fixed to 256.
For performance evaluation, we strictly follow the ADM setup~\cite{dhariwal2021diffusion} and adopt several popular metrics: Fréchet Inception Distance (FID)~\cite{fid}, structural FID (sFID)~\cite{sfid}, Inception Score (IS)~\cite{is}, Precision (Prec.) and Recall (Rec.)~\cite{precrecall}, all calculated from 50K generated images.
Classifier-free guidance (CFG)~\cite{cfg} is not employed unless specified.
We use Euler-Maruyama sampler with 250 steps for sampling.
All experiments are conducted on 8 $\times$ 80GB H800 GPUs.
More implementation details are provided in \cref{sec:supp:implementation_details}.

\noindent \textbf{Compared baselines}
For multi-step comparison, we compare our method against existing alternatives from three categories:
1) pixel-diffusion: ADM~\cite{dhariwal2021diffusion}, VDM++~\cite{kingma2023understanding}, CDM~\cite{ho2022cascaded};
2) latent-diffusion with UNet: LDM~\cite{rombach2022high};
3) diffusion transformers: SiT~\cite{sit}, DiT~\cite{dit}, SD-DiT~\cite{zhu2024sd}, MaskDiT~\cite{zheng2023fast}, MDTv2~\cite{gao2023mdtv2};
and 4) current state-of-the-art models: REPA~\cite{repa}, REG~\cite{wu2025representation}, E2E-REPA~\cite{leng2025repa}, SRA~\cite{jiang2025no}, DispLoss~\cite{wang2025diffuse}. 
For one-step comparison, we compare our method with recent popular methods:
iCT-XL/2~\cite{ict}, Shortcut-XL/2~\cite{shortcut}, IMM-XL/2~\cite{inductive} and Meanflow~\cite{geng2025mean}.
Detailed descriptions of these methods are given in~\cref{sec:supp:comparison_baselines}.

\subsection{Main Results}
In this part, we first present comparison results of applying our methods on different baselines across various scales to investigate its effectiveness and scalability, and then compare our methods against existing SoTA models.

\begin{figure}[t]
    \begin{center}  \centerline{\includegraphics[width=\linewidth]{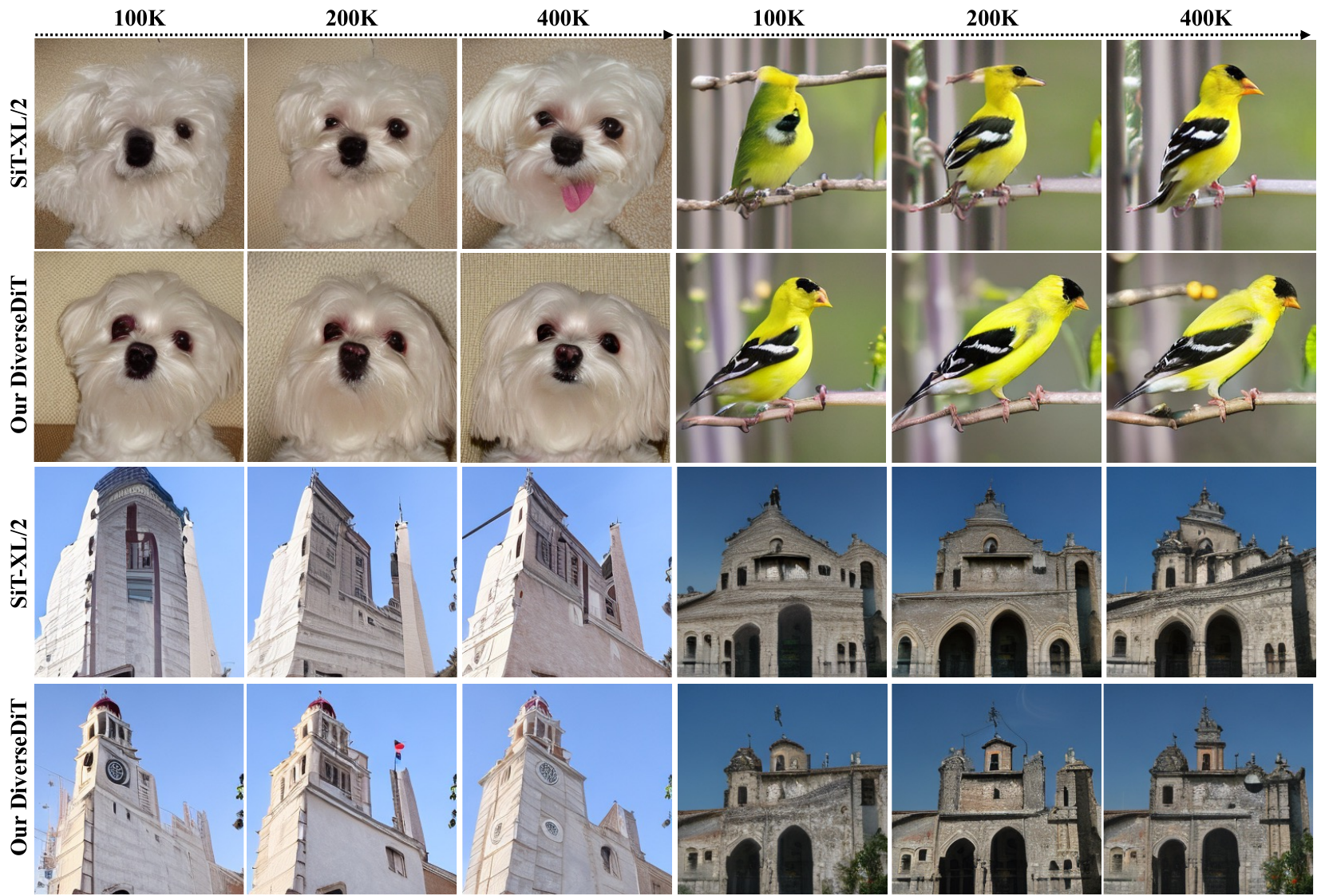}}
    \vskip -0.1in
    \caption{\textbf{Generated samples from different training iterations.} 
    %
    Images are sampled using the same seed, noise and class label. 
    We use a classifier-free guidance scale of 4.0 during sampling.
    }
    \label{fig:iter_qualitative}
    \end{center}
    \vskip -0.4in
\end{figure}

\begin{table}[]
\centering
\scriptsize
\caption{\textbf{Comparison results on ImageNet $256\times256$ with CFG}.}
\vskip -0.1in
\setlength{\tabcolsep}{2mm}{
\begin{tabular}{llccccc}
\toprule
\textbf{Method}     & \textbf{Epochs}  & \textbf{FID$_\downarrow$}   & \textbf{sFID$_\downarrow$} & \textbf{IS$_\uparrow$} & \textbf{Pre.$_\uparrow$} & \textbf{Rec.$_\uparrow$} \\ \midrule
ADM-U~\cite{dhariwal2021diffusion} 
& 400       & 3.94     & 6.14       & 186.70    & 0.82    & 0.52  \\
VDM++~\cite{kingma2023understanding}
& 560       & 2.40     & -          & 225.30    & -       & -     \\
CDM~\cite{ho2022cascaded}
& 2160      & 4.88     & -          & 211.80    & -       & -     \\
LDM-4~\cite{rombach2022high}
& 200       & 3.60     & -          & 247.70    & \textbf{0.87}    & 0.48  \\
SD-DiT~\cite{zhu2024sd}
& 480       & 3.23     & -          & -         & -       & -     \\
MakDiT~\cite{zheng2023fast}
& 1600      & 2.28     & 5.67       & 276.70    & 0.80    & 0.62  \\
MDTv2-XL/2~\cite{gao2023mdtv2} 
& 1080      & 1.58     & 4.52       & \textbf{314.70}    & 0.79    & 0.65  \\
DiT-XL/2~\cite{dit}
& 1400      & 2.27     & 4.60       & 278.20    & 0.83    & 0.57  \\ 
SiT-XL/2~\cite{sit}
& 1400      & 2.06     & 4.50       & 270.30    & 0.82    & 0.59  \\
REPA~\cite{repa}
& 200       & 1.96     & 4.49       & 264.00    & 0.82    & 0.60  \\
REPA~\cite{repa}
& 800       & 1.80     & 4.50       & 284.00    & 0.81    & 0.61  \\
REG~\cite{wu2025representation}
& 800       & \textbf{1.36}     & 4.25       & 299.40    & 0.77    & 0.66  \\
E2E-REPA~\cite{leng2025repa}
& 800       & 1.69     & \textbf{4.17}       & 219.30    & 0.77    & \textbf{0.67}  \\
SRA~\cite{jiang2025no}
& 800       & 1.58     & 4.65       & 311.40    & 0.80    & 0.63  \\
DispLoss~\cite{wang2025diffuse}
& 800       & 1.97     & -          & -         & -       & -     \\
\midrule
\textbf{DiverseDiT (Ours)}
& 80        & 1.89     & 4.44       & 276.85    & 0.81    & 0.66  \\
\textbf{DiverseDiT (Ours)}
& 200       & 1.52     & 4.23       & 282.72    & 0.81    & 0.66  \\ 
\bottomrule
\end{tabular}
}
\label{tab:res256}
\vskip -0.1in
\end{table}

\begin{table}[]
\vspace{-0.1cm}
\centering
\scriptsize
\caption{\textbf{Comparison results on ImageNet $512\times512$ with CFG}.}
\vskip -0.1in
\setlength{\tabcolsep}{2mm}{
\begin{tabular}{llccccc}
\toprule
\textbf{Method}     & \textbf{Epochs}  & \textbf{FID$_\downarrow$}   & \textbf{sFID$_\downarrow$} & \textbf{IS$_\uparrow$} & \textbf{Pre.$_\uparrow$} & \textbf{Rec.$_\uparrow$} \\ \midrule
ADM-G~\cite{dhariwal2021diffusion} 
& 400       & 2.85     & 5.86       & 221.70    & {\bf0.84}    & 0.53  \\
VDM++~\cite{kingma2023understanding}
& -         & 2.65     & -          & \bf{278.10} & -       & -     \\
MakDiT~\cite{zheng2023fast}
& 800       & 2.50     & 5.10       & 256.30    & 0.83    & 0.56  \\
DiT-XL/2~\cite{dit}
& 600       & 3.04     & 5.02       & 240.80    & {\bf0.84}    & 0.54  \\ 
SiT-XL/2~\cite{sit}
& 600       & 2.62     & {\bf4.18}  & 252.20    & {\bf0.84}    & 0.57  \\
REPA~\cite{repa}
& 200       & 2.08     & 4.19       & 274.60    & 0.83    & 0.58  \\
\midrule
\textbf{DiverseDiT (Ours)}
& 80        & 2.21     & 4.31       & 241.08    & 0.82    & {\bf0.61}  \\
\textbf{DiverseDiT (Ours)}
& 200       & {\bf1.99}& {\bf4.19}  & 267.12    & 0.83    & {\bf0.61}  \\ 
\bottomrule
\end{tabular}
}
\label{tab:res512}
\vskip -0.3in
\end{table}

\noindent \textbf{Improving representation learning across various model scales.}
\cref{tab:different_size} presents the quantitative results of applying our proposed techniques to SiT and REPA across various model scales on ImageNet 256$\times$256 without CFG.
We could observe that incorporating our method consistently yields substantial improvements on all evaluation metrics across all model scales, demonstrating its effectiveness and generalization regardless of the underlying training paradigm.
Notably, our method achieves an FID of 17.29 on the REPA-B setting with 400K iterations, which is better than that of SiT-L (\emph{i.e.,} 18.77) with the same training iterations.
Similarly, the performance of applying our method on the REPA-L outperforms that of the REPA-XL, \emph{i.e.,} 8.47 vs 8.73 on FID and 123.03 vs 118.68 on IS.
Moreover, \cref{fig:iter_qualitative} shows the images generated by SiT-XL and our proposed method at different training iterations.
Our generated images exhibit more details, better structures, and fewer artifacts, demonstrating that our method leads to faster convergence and higher visual quality compared to the baseline models.
That is, our design towards improving the diversity of representations across different blocks contributes to a scalable and efficient learning process.

\begin{figure*}[t]
    \vskip -0.1in
    \begin{center}  \centerline{\includegraphics[width=.99\linewidth]{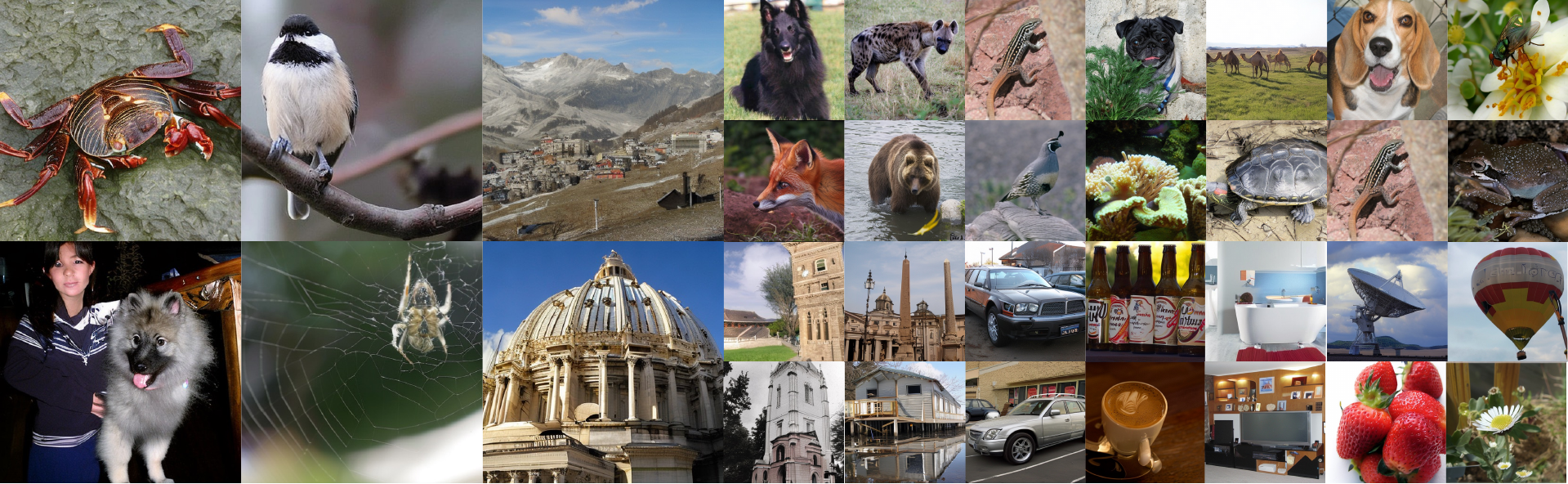}}
    \vskip -0.05in
    \caption{\textbf{Generated samples on ImageNet 256$\times$256 from our DiverseDiT}.
    We use a classifier-free guidance scale of 4.0.
    %
    }
    \label{fig:qualitative}
    \end{center}
    \vskip -0.4in
\end{figure*}

\noindent \textbf{Comparison with SoTA Models.}
\cref{tab:res256} presents the comparison results with recent state-of-the-art (SoTA) methods using CFG On ImageNet $256\times256$. 
As can be observed from the table, our method achieves competitive performance compared to SoTA models while requiring significantly fewer training epochs.
At 80 epochs, our method attains an FID score of 1.89, outperforming REPA trained for 200 epochs (1.96) and surpassing the performance of several established methods trained for hundreds or even thousands of epochs.
For instance, the SiT-XL/2 model requires 1400 epochs to reach an FID of 2.06, while we achieve 1.52 with only 200 epochs.
While REG achieves a slightly better FID of 1.36, it requires 800 epochs, four times the training cost of ours. 
Furthermore, \cref{tab:res512} shows the comparison results on ImageNet $512\times512$.
Similarly, our DiverseDiT achieves a comparable FID of 2.21 with only 80 epochs and obtains the best FID score when trained for 200 epochs.
The consistent strong performance across multiple metrics, coupled with the significantly reduced training time, shows the efficiency and effectiveness of our model in learning diverse and high-quality representations.
Additionally, we provide selected samples generated by our method in \cref{fig:qualitative}, the generated images demonstrate that DiverseDiT produces images with excellent quality.
More quantitative and qualitative results CFG are given in \cref{sec:supp:more_quantitative} and \cref{sec:supp:more_qualitative_results}.

\noindent \textbf{Improving representation learning for one-step generation.}
Regarding one-step generation, we applied our proposed techniques to MeanFlow (MF)~\cite{geng2025mean} to assess the generalization ability.
\cref{tab:different_size_mf} presents the quantitative results across different model scales.
Similar to our previous findings, incorporating our method consistently improves the performance with different model sizes, \emph{e.g.,} our method improves the FID score of MF-B/2 from 9.44 to 8.51 and the IS from 152.55 to 158.84.
Additionally, \cref{tab:mf} shows a comparison of our method with other one-step generative models.
Notably, we achieve a new SoTA with an FID score of 2.99 by applying our method to MeanFlow-XL/2.
These results identify the effectiveness of our method in improving representation learning for one-step generation.

\begin{table}[t]
    \centering
    \vskip -0.1in
    \caption{\textbf{Variation in model scale on ImageNet 256$\times$256 for one-step generation without CFG}. 
    %
    }
    \scriptsize
    \vskip -0.1in
    \setlength{\tabcolsep}{1.8mm}{
    \begin{tabular}{lccccccc}
        \toprule
        \textbf{Model}  & \textbf{Iter}. & \textbf{Step}. & \textbf{FID}$_\downarrow$ & \textbf{sFID}$_\downarrow$ & \textbf{IS}$_\uparrow$ & \textbf{Prec.}$_\uparrow$ & \textbf{Rec.}$_\uparrow$ \\
        \midrule
        MF-B/2            & 400K & 1 & 9.44          & 6.21          & 152.55          & 0.77          & 0.42          \\
        \textbf{+ (Ours)} & 400K & 1 & \textbf{8.51} & \textbf{5.95} & \textbf{158.84} & \textbf{0.78} & \textbf{0.44} \\
        \midrule
        MF-L/2            & 400k & 1 & 8.73          & 6.11          & 161.69          & 0.79          & 0.40          \\
        \textbf{+ (Ours)} & 400k & 1 & \textbf{7.15} & \textbf{5.91} & \textbf{199.66} & \textbf{0.81} & \textbf{0.41} \\
        \midrule
        MF-XL/2           & 400k & 1 & 5.94          & 6.10          & 213.13          & \textbf{0.83} & \textbf{0.41} \\
        \textbf{+ (Ours)} & 400k & 1 & \textbf{5.69} & \textbf{5.88} & \textbf{214.92} & \textbf{0.83} & \textbf{0.41} \\
        \bottomrule
    \end{tabular}
    }
    \vskip -0.1in
    \label{tab:different_size_mf}
\end{table}

\begin{table}[t]
  \small
  \centering
  \caption{\textbf{Comparison results on ImageNet 256$\times$256 for one-step generation with CFG}.}
  \vskip -0.1in
  \setlength{\tabcolsep}{1.8mm}{
  \begin{tabular}{lcccr}
    \toprule
    \textbf{Method} & \textbf{Params} & \textbf{Step} & \textbf{NFE} & \textbf{FID$_\downarrow$} \\ \midrule
    iCT-XL/2 \cite{ict}                           & 675M & 1 & 1 & 34.24 \\
    Shortcut-XL/2 \cite{shortcut}                 & 675M & 1 & 1 & 10.60 \\
    IMM-XL/2 \cite{inductive}                     & 675M & 1 & 2 & 7.77 \\
    MF-XL/2 \cite{geng2025mean}                   & 676M & 1 & 1 & 3.43 \\
    MF-XL/2 + DispLoss~\cite{wang2025diffuse}     & 676M & 1 & 1 & 3.21 \\
    MF-XL/2 + \textbf{ours}                       & 713M & 1 & 1 & \textbf{2.99} \\
    \bottomrule
  \end{tabular}
  }
  \vskip -0.2in
\label{tab:mf}
\end{table}

\subsection{Ablation Analysis}
\label{sec:ablation}
In this part, we perform ablation studies to testify the efficacy of each component and the design choices used in our main experiments.

\noindent \textbf{Ablation on designed components.}
In \cref{tab:ablation_components}, we present an ablation study that analyzes the contribution of different components of our method by removing each component.
The results clearly demonstrate the importance of both the representation diversity loss and the long residual connections for optimal performance. 
Removing the diversity loss (w/o diversity) worsens FID scores for both SiT-B (from 28.05 to 32.77) and REPA-B (from 17.29 to 20.66). 
Similarly, removing the long residual connections (w/o residual) also noticeably increases FID for both baseline models. 
These results confirm that both components of DiverseDiT play a crucial role in promoting diverse representation learning and improving the model performance.
More ablation analysis results are presented in \cref{sec:supp:more_ablation}.

\begin{table}[t]
    \centering
    \small
    \caption{\textbf{Ablation analysis on different components.}}
    \label{tab:ablation_components}
    \vskip -0.1in
    \setlength{\tabcolsep}{2mm}{
    \begin{tabular}{lccccc}
        \toprule
         \textbf{Component} & \textbf{FID}$_\downarrow$ & \textbf{sFID}$_\downarrow$ & \textbf{IS}$_\uparrow$ & \textbf{Prec.}$_\uparrow$ & \textbf{Rec.}$_\uparrow$ \\
         \midrule
        SiT-B + ${\texttt{full}}$    & 28.05         & 6.04          & 50.66         & 0.57          & 0.63         \\
        w/o ${\texttt{diversity}}$   & 32.77         & 6.42          & 44.85         & 0.54          & 0.63         \\
        w/o ${\texttt{residual}}$    & 33.72         & 6.53          & 43.97         & 0.53          & 0.63         \\ \midrule
        REPA-B + ${\texttt{full}}$   & 17.29         & 6.56          & 79.92         & 0.62          & 0.65        \\
        w/o ${\texttt{diversity}}$   & 20.66         & 6.66          & 72.72         & 0.61          & 0.64         \\
        w/o ${\texttt{residual}}$    & 18.18         & 6.62          & 75.49         & 0.61          & 0.65         \\
        \bottomrule
    \end{tabular}
    }
    \vskip -0.2in
\end{table}

\noindent \textbf{Effect of diversity loss variants.}
In the main experiments, we adopt a combination of different discrepancies for the representation diversity loss. 
This part validates the contribution of different components of the diversity loss in \cref{tab:ablation_loss}.
The results show that using all components of the diversity loss (REPA-B + full) achieves the best performance (FID 17.29, IS 79.92). 
Moreover, removing any single loss component degrades performance.
More importantly, adding any component consistently outperforms the REPA-B baseline (\cref{tab:different_size}), demonstrating their effectiveness in encouraging diverse representation learning.
%
%
%

\begin{table}[t]
    \centering
    \small
    \caption{\textbf{Ablation analysis on different loss variants.}}
    \label{tab:ablation_loss}
    \vskip -0.1in
    \setlength{\tabcolsep}{2mm}{
    \begin{tabular}{lccccc}
        \toprule
         \textbf{Component} & \textbf{FID}$_\downarrow$ & \textbf{sFID}$_\downarrow$ & \textbf{IS}$_\uparrow$ & \textbf{Prec.}$_\uparrow$ & \textbf{Rec.}$_\uparrow$ \\
         \midrule
         REPA-B + ${\texttt{full}}$                     & 17.29   & 6.56         & 79.92          & 0.62          & 0.65         \\
         ${\texttt{only $\mathcal{L}_{\text{orth}}$}}$  & 18.97   & 6.64         & 75.44          & 0.61          & 0.65         \\
         ${\texttt{only $\mathcal{L}_{\text{MI}}$}}$    & 17.70   & 6.62         & 78.34          & 0.62          & 0.65         \\
         ${\texttt{only $\mathcal{L}_{\text{div}}$}}$   & 20.85   & 6.78         & 68.74          & 0.61          & 0.65         \\
        \bottomrule
    \end{tabular}
    }
    \vskip -0.2in
\end{table}

\noindent \textbf{Effect of the adaptive range of diversity loss.}
In \cref{tab:ablation_range}, we test the impact of different adaptive ranges of our diversity loss on the performance.
A constant weight leads to divergence, possibly due to excessive representation discrepancy as discussed in \cref{sec:method}.
By contrast, adaptively controlling the diversity loss based on the loss value enables stable training.
In particular, $[0.1, 0.5]$ yields the best performance, suggesting that a narrower adaptive range effectively promotes diversity without sacrificing image quality.

\begin{table}[t]
    \centering
    \small
    \caption{\textbf{Ablation study on adaptive range of diversity loss.}}
    \label{tab:ablation_range}
    \vskip -0.1in
    \setlength{\tabcolsep}{2mm}{
    \begin{tabular}{lccccc}
        \toprule
         \textbf{Range} & \textbf{FID}$_\downarrow$ & \textbf{sFID}$_\downarrow$ & \textbf{IS}$_\uparrow$ & \textbf{Prec.}$_\uparrow$ & \textbf{Rec.}$_\uparrow$ \\
         \midrule
         ${\texttt{Adap}}$ $[0.1, 0.5]$    & 28.05         & 6.04          & 50.66         & 0.57          & 0.63        \\
         ${\texttt{Adap}}$ $[0.2, 0.7]$    & 30.59         & 6.34          & 48.07         & 0.55          & 0.63        \\
         ${\texttt{Adap}}$ $[0.3, 0.9]$    & 31.85         & 6.43          & 45.98         & 0.54          & 0.63        \\ 
         ${\texttt{Constant}}$             & \texttt{diverge} & -             & -             & -             & -        \\
        \bottomrule
    \end{tabular}
    }
    \vskip -0.15in
\end{table}

\noindent \textbf{Combining with existing methods for further improvement.}
\cref{tab:compatibility} shows \ourmethod's compatibility with DispLoss~\cite{wang2025diffuse} and SRA~\cite{jiang2025no}, yielding further performance gains.
The results demonstrate that our method can be effectively combined with existing approaches for further performance improvements, reflecting the flexibility of our approach.
Noticeably, when combining our proposed method with both DispLoss and SRA, we achieve an FID of 21.95, which is better than that of REPA (22.99 in \cref{tab:different_size}) at the same iterations.
Recall that REPA requires external models for representation alignment, while here we do not rely on any external guidance, demonstrating the potential for representation learning through internal mechanisms.

\begin{table}[t]
    \centering
    \small
    \caption{
    \textbf{Combining our method with prior approaches.}
    All results are calculated from 400K iterations without CFG.
    }
    \label{tab:compatibility}
    \vskip -0.1in
    \setlength{\tabcolsep}{1.8mm}{
    \begin{tabular}{lccccc}
        \toprule
         \textbf{Component} & \textbf{FID}$_\downarrow$ & \textbf{sFID}$_\downarrow$ & \textbf{IS}$_\uparrow$ & \textbf{Prec.}$_\uparrow$ & \textbf{Rec.}$_\uparrow$ \\
         \midrule
         SiT-B                      & 36.80         & 6.77         & 40.09          & 0.51          & 0.63         \\
         + ${\texttt{Ours}}$        & 28.05         & 6.04         & 50.66          & 0.57          & 0.63         \\
         ++ ${\texttt{DispLoss}}$~\cite{wang2025diffuse}   & 24.98         & 6.01         & 57.04          & 0.59          & 0.63         \\
         +++ ${\texttt{SRA}}$~\cite{jiang2025no}        & 21.95         & 5.92         & 64.64          & 0.60          & 0.64         \\
        \bottomrule
    \end{tabular}
    }
    \vskip -0.2in
\end{table}
\section{Related Works}
\label{sec:related}

\noindent \textbf{Diffusion Models.}
Diffusion probabilistic models~\cite{ddpm,sohl2015deep,ddim_song2020}, which generate images via iteratively denoising Gaussian noises, have become the dominating paradigm for image~\cite{rombach2022high,chen2023pixart,imagen3_baldridge2024} and video generation~\cite{yang2024cogvideox,hunyuan_video_2024,gupta2024photorealistic}, driven by improved training stability with flow matching~\cite{flowmatching_lipman2022,RectifiedFlow_liuflow} and exceptional model scalability from conventional UNet-based models~\cite{rombach2022high,dhariwal2021diffusion,blattmann2023stable} to the Transformer-based architectures~\cite{dit,sit,bao2023all,attention_vaswani2017}.
Besides these advancements, many efforts improve diffusion models from the perspective of accelerating sampling process~\cite{CM_song2023consistency,geng2025mean,dpmsolver_lu2022}, designing noise schedules~\cite{karras2022elucidating,lu2024simplifying,nichol2021improved}, developing novel architectures like linear transformers~\cite{xie2025sana,chen2025sana}, MoE-based models~\cite{shi2025diffmoe,wei2025routing}, \emph{etc.}

\noindent \textbf{Representation Learning.}
Central to representation learning is to learn rich and meaningful representations for downstream tasks.
This field has evolved through several key paradigms, including discriminative, generative, and multimodal approaches.
Discriminative methods, exemplified by contrastive learning based methods including BYOL~\cite{grill2020bootstrap}, DINO~\cite{caron2021emerging,oquab2023dinov2,simeoni2025dinov3}, and MoCo~\cite{he2020momentum,chenxinlei2020improved,chen2021empirical}, capture discriminative signals between images to learn strong representations.
The generative variant learns the underlying data distribution via reconstructing input images, representative works including auto-encoder methods VAE~\cite{kingma2013auto}, MAE~\cite{he2022masked}, and masked image modeling~\cite{xie2022simmim}.
Similarly, diffusion models also learn informative features as inherent denoising autoencoders~\cite{yang2023diffusion,mittal2023diffusion,zhang2022unsupervised}.
To enable cross-model understanding and retrieval, multimodal methods~\cite{CLIP_radford2021learning,li2022blip,zhai2023sigmoid,tschannen2025siglip} align textual and visual signals in a shared representation space.
Despite these advancements, it remains unclear what representations should be learned for different tasks, especially for diffusion generative models.

\noindent \textbf{Representation Learning of Diffusion Models.}
Many prior approaches identify that improved representation learning of diffusion models advances both synthesis quality and downstream tasks~\cite{li2023dreamteacher,xiang2023denoising}.
Following this philosophy, REPA~\cite{repa} incorporated external encoders to align diffusion representations with pre-trained representations, extended by REPA-E~\cite{leng2025repa} that enabled end-to-end training with VAE and SARA~\cite{chen2025sara} that introduced structural alignment.
Further, SoftREPA~\cite{softrepa} performed alignment on textual embeddings and REG~\cite{wu2025representation} entangled image latents and class tokens  to harness discriminative representations.
Unlike these approaches that requires external guidances for knowledge alignment, SRA~\cite{jiang2025no} leveraged representations from later layers with lower noise to guide representations of earlier layers with higher noise.
Further, Wang \etal~\cite{wang2025diffuse} developed DispLoss to regularize internal features to disperse in the embedding space, thus encouraging the model to capture informative representations.
Despite these advancements, it remains unclear how meaningful representations are learned and what representations within models are more suitable.
We thus perform a systematic investigation on this and develop \ourmethod to learn diverse and effective representations for diffusion transformers.
\section{Conclusion}
\label{sec:conc}

In this work, we first present a comprehensive analysis of the representation learning process within diffusion transformers (DiTs), revealing the critical role of representation diversity across different blocks.
Based on these insights, we introduce \ourmethod, a novel and efficient framework explicitly designed to enhance representation diversity with long residual connections to diversify input and representation diversity loss to encourage distinct features across blocks, without relying on external guidance.
Extensive experiments demonstrate that \ourmethod consistently improves training convergence and model performance across various model scales and settings, including multi-step and one-step generation.
These findings contribute to a deeper understanding of representation learning dynamics in DiTs and offer a practical, effective strategy for boosting their performance, paving the way for future research in learning representations for generative models.

\subsubsection*{Acknowledgments}
This work was supported by AI for Science Program, Shanghai Municipal Commission of Economy and Informatization (Grant No. 2025-GZL-RGZN-BTBX-02017).

{
    \small
    \bibliographystyle{misk/ieeenat_fullname}
    \bibliography{ref}
}

\clearpage
\setcounter{page}{1}
\maketitlesupplementary

\appendix
\newcommand{\AppendixPrefix}{A}
\renewcommand{\thefigure}{\AppendixPrefix\arabic{figure}}
\setcounter{figure}{0}
\renewcommand{\thetable}{\AppendixPrefix\arabic{table}} 
\setcounter{table}{0}
\renewcommand{\theequation}{\AppendixPrefix\arabic{equation}} 
\setcounter{equation}{0}

\section{Appendix Overview}
This supplementary material is organized as follows:
First, we present more implementation details in~\cref{sec:supp:implementation_details}.
Followed by the evaluation details and brief introduction of comparison baseline methods in~\cref{sec:supp:eval} and~\cref{sec:supp:comparison_baselines}, respectively.
Then, \cref{sec:supp:analysis_details} shows the detailed quantitative results of our comprehensive analysis in~\cref{sec:revisting_repa}.
In the following, we present more quantitative comparison results under various settings in~\cref{sec:supp:more_quantitative} and more ablation results in~\cref{sec:supp:more_ablation}.
Moreover, \cref{sec:supp:limitations} discusses the limitations and potential future works of our method.
Finally, \cref{sec:supp:more_qualitative_results} illustrates more uncurated images generated by our proposed method.

\begin{table*}[h!]
    \centering
    \small
    \caption{
    \textbf{Hyperparameter setup for the main experiment.}
    We strictly follow the setup of the baseline SiT and REPA for our training and evaluation for fair comparison.
    }
    \begin{tabular}{l c c c c c}
        \toprule
        & Table 1 (SiT-B) & Table 1 (SiT-L) & Table 1 (SiT-XL) & Table 2 (SiT-XL) & Table 3 (SiT-XL) \\
        \midrule
        \textbf{SiT + Ours} \\
        Input dim. & 32$\times$32$\times$4 & 32$\times$32$\times$4 & 32$\times$32$\times$4 & 32$\times$32$\times$4 & 64$\times$64$\times$4 \\
        Num. layers & 12 & 24 & 28 & 28 & 28 \\
        Hidden dim. & 768 & 1,024 & 1,152 & 1,152 & 1,152 \\
        Num. heads & 12 & 16 & 16 & 16 & 16 \\ 
        \midrule
        \textbf{REPA + Ours} \\
        $\lambda$       & 0.5 & 0.5 & 0.5 & 0.5 & 0.5 \\
        Alignment depth & 5   & 8 & 8 & 8 & 8 \\
        $\mathrm{sim}(\cdot, \cdot)$ & cos. sim.  & cos. sim. & cos. sim & cos. sim.  & cos. sim.  \\
        Encoder $f(x)$ & DINOv2-B  & DINOv2-B & DINOv2-B & DINOv2-B  & DINOv2-B\\
        \midrule
        \textbf{Optimization} \\
        Loss adaptive range & [0.1, 0.5] & [0.1, 0.5] & [0.1, 0.5] & [0.1, 0.5]  & [0.1, 0.5] \\
        Training iteration   & 400K     & 400K & 400K & 4M & 1M \\ 
        Training Batch size  & 256      & 256 & 256 & 256 & 256 \\ 
        Optimizer            & AdamW & AdamW & AdamW & AdamW & AdamW \\
        lr                   & 0.0001 & 0.0001 &  0.0001 & 0.0001 & 0.0001 \\
        $(\beta_1, \beta_2)$ & (0.9, 0.999) & (0.9, 0.999) & (0.9, 0.999) & (0.9, 0.999) & (0.9, 0.999) \\
        \midrule
        \textbf{Interpolants} \\
        $\alpha_t$ & $1-t$ & $1-t$ & $1-t$ & $1-t$ & $1-t$ \\
        $\sigma_t$ & $t$ & $t$ & $t$ & $t$ & $t$ \\
        $w_t$ & $\sigma_t$ & $\sigma_t$ & $\sigma_t$ & $\sigma_t$ & $\sigma_t$ \\
        Training objective & v-prediction & v-prediction & v-prediction & v-prediction & v-prediction \\
        Sampler & Euler-Maruyama & Euler-Maruyama & Euler-Maruyama & Euler-Maruyama & Euler-Maruyama \\
        Sampling steps & 250 & 250 & 250 & 250 & 250 \\
        Guidance       & 1.0 & 1.0 & 1.0 & 1.35 &  1.35 \\
        \bottomrule
    \end{tabular}
    \vskip -0.2in
    \label{tab:supp_hyperparam}
\end{table*}

\section{More Implementation Details}
\label{sec:supp:implementation_details}

\begin{figure}[t]
    \begin{center}  \centerline{\includegraphics[width=\linewidth]{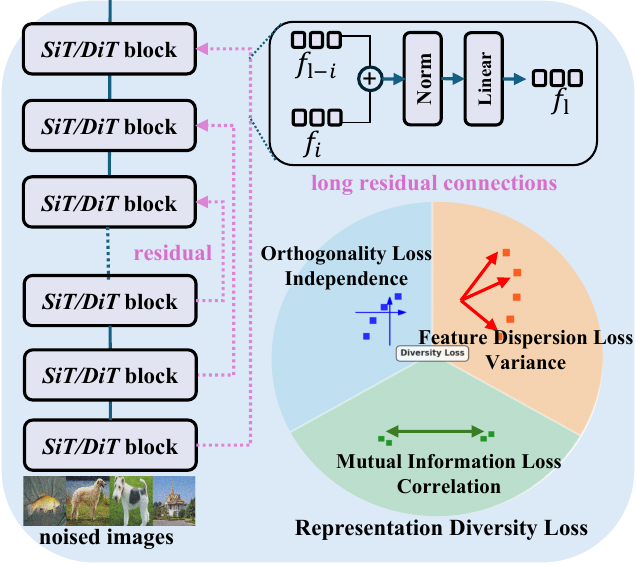}}
    \caption{\textbf{Detailed diagram of our proposed DiverseDiT.} 
    \ourmethod incorporates long residual connections to diversify input representations across blocks and a representation diversity loss to encourage blocks to learn distinct features.
    }
    \label{fig:method_supp}
    \end{center}
    \vskip -0.2in
\end{figure}

\paragraph{Detailed diagram of our DiverseDiT}
Our diversity loss is theoretically motivated from the following perspectives:
1) introducing an explicit inductive bias that encourages block-wise diversity to model the underlying observed distribution;
2) improving the representational orthogonality across blocks, thus reducing the redundancy and mutual correlation of different blocks;
3) promoting the coverage of the representation space, enabling blocks to specialize in complementary structures.
For our proposed long residual connections and representation diversity loss, we first concatenate the hidden features of two blocks and perform layer normalization and a lightweight linear layer, as illustrated in \cref{fig:method_supp}.

\paragraph{More implementation details}
We implement our proposed techniques on the original SiT~\cite{sit} and REPA~\cite{repa} implementation, leaving other details unchanged.
Regarding the representation diversity loss, we calculate the corresponding loss items following the definition of each loss in \cref{sec:method}, we randomly select $10$ layers for computing the loss, as we find that computing on more layers gains similar performance improvement. 
Such observation aligns with that of DispLoss~\cite{wang2025diffuse}, where the effect of representation diversity loss propagates to other blocks, even though it is not directly applied to them.
Further, the detailed setup for hyperparameters is presented in \cref{tab:supp_hyperparam}.
To speed up the training process and save pre-processing time, we adopt mixed-precision (fp16) with gradient clipping and pre-compute VAE latents with stable diffusion VAE~\cite{rombach2022high} (sd-vae-ft-mse) following REPA.
For all optimization, we adopt AdamW~\cite{loshchilov2017decoupled} with a learning rate of 1e-4, and the training batchsize is 256.
When classifier-free guidance (CFG)~\cite{cfg} is applied for generating images, we use the same guidance interval as that of REPA, which has been identified to improve the model performance, also illustrated in our results in \cref{{tab:supp_different_depth_cfg}}.
Additionally, we implement our method on the official MeanFlow~\cite{geng2025mean} and follow their default configurations for training and evaluation for one-step generation experimental evaluation.

\paragraph{Sampler}
Following the practice of REPA, we employ the Euler-Maruyama sampler with the SDE sampling with a diffusion coefficient $\sigma_t$. The sampling step for generating each image is set to 250.

\paragraph{Discussion on hyperparameter sensitivity}
In our evaluation, we use the same hyperparameter settings when applying our method to various models, namely SiT~\cite{sit}, REPA~\cite{repa}, DispLoss~\cite{wang2025diffuse}, SRA~\cite{jiang2025no} and MeanFlow~\cite{geng2025mean}.
Our method consistently improves performance across different backbones and different sampler settings (multi-step and one-step sampling).
These results indicate that although our diversity loss introduces several hyperparameters, it is robust to hyperparameter choices.

\paragraph{Computing resources}
All experiments were conducted on NVIDIA H100 (80GB) or H200 (141GB) GPUs.
For training, the speed is about 5.8 steps/s for training SiT-XL + Ours, and it takes about 1.38 hours to generate 50,000 images for evaluation (10.04 images/s).
We have uploaded the compute report for detailed GPU hours used for our analysis and evaluation.

\paragraph{Pretrained vision foundation models for representation alignment}
In our systematic analysis and experimental evaluation, we use three pretrained visual encoders as external representation guidance, namely DINOv2-B~\cite{oquab2023dinov2}, MAE~\cite{he2022masked}, and MoCov3~\cite{chen2021empirical}.
For DINOv2-B\footnote{\url{https://github.com/facebookresearch/dino}} and MAE\footnote{\url{https://github.com/facebookresearch/mae}} models, we download the pretrained model officially released by the original authors.
Regarding the MoCov3 model, we download the -L version from the implementation of RCG\footnote{\url{https://github.com/LTH14/rcg}}~\cite{li2023return}; following REPA.
For representation alignment, we perform projection with three MLP layers with SiLU activations following the exact configuration of REPA.
\begin{itemize}[leftmargin=0.2in]
\item \textbf{DINOv2}~\cite{oquab2023dinov2}: DINOv2 employs a vision transformer (ViT) architecture and learns self-supervised representations by enforcing consistency between different views of an image. It measures the feature distance between the representations of real and generated images, capturing high-level semantic information.
\item \textbf{MAE}~\cite{he2022masked}: MAE trains an encoder and a lightweight decoder with a reconstruction objective. It learns to reconstruct masked patches of an image, learning robust representations.
\item \textbf{MoCov3}~\cite{chen2021empirical}: based on the philosophy of contrastive learning, MoCov3 empirically revisits prior MoCo series~\cite{he2020momentum, chenxinlei2020improved} and scales to larger model sizes to learn representations by maximizing the similarity between different views of the same image while minimizing the similarity between views of different images.
\end{itemize}

\section{Evaluation Details}
\label{sec:supp:eval}

\paragraph{Implementation details for evaluation}
We strictly follow the setup and use the same reference batches of ADM~\cite{dhariwal2021diffusion} for evaluation, following their official implementation.
Specifically, for 256$\times$256 evaluation, we generate 50,000 images and convert them into a .npz file and compute the quantitative metrics from the reference batch (VIRTUAL\_imagenet256\_labeled.npz) of ADM\footnote{\url{https://github.com/openai/guided-diffusion/tree/main/evaluations}}.
Similarly, we quantify the result of 512$\times$512 evaluation via computing the metrics between our generated images and the reference batch (VIRTUAL\_imagenet512.npz).

\paragraph{Evaluation metrics}
We adopt several popular metrics for evaluation: Fréchet Inception Distance (FID)~\cite{fid}, structural FID (sFID)~\cite{sfid}, Inception Score (IS)~\cite{is}, Precision (Prec.) and Recall (Rec.)~\cite{precrecall}. Their main concepts are:
\begin{itemize}[leftmargin=0.2in]
\item \textbf{FID}~\cite{fid} computes the Fr\'{e}chet Distance between two observed data distributions, which represent the feature distributions of synthesized and real images extracted by the pre-trained Inception-V3~\cite{szegedy2016rethinking}.
Formally,  FID is calculated by
\begin{align}
\mathrm{FID(X,Y)}=\left\|\mu_s-\mu_r\right\|^2+\operatorname{Tr}\left(\Sigma_s+\Sigma_r-2\left(\Sigma_s \Sigma_r\right)^{\frac{1}{2}}\right),
\end{align}
where $\mathrm{X}$ and $\mathrm{Y}$ represent the synthesized distribution and real  distribution, respectively.
$\mu$ and $\Sigma$ correspond to the mean and variance of the distribution, and $\operatorname{Tr(\cdot)}$ is the trace operation.
\item \textbf{sFID}~\cite{sfid} is a variant of FID that aims to be more robust to structural differences between real and generated images. Instead of using the standard Inception-V3 features, sFID uses features extracted from different layers of the network, focusing on structural information. This makes it more sensitive to the arrangement of objects and their parts, and less sensitive to color or texture differences.
\item \textbf{IS}~\cite{is} measures the quality and diversity of generated images. It uses the Inception-V3 model to predict the class of each generated image. A good Inception Score means that the generated images are clear and belong to a specific class (high confidence), and that the generated images cover a wide range of classes (high diversity).
Formally, Inception Score is calculated by:
\begin{align}
\mathrm{IS} = \exp(\mathbb{E}_{\mathrm{x} \sim \mathrm{X}}[D_{KL}(p(\mathrm{y}|\mathrm{x}) || p(\mathrm{y}))])
\end{align}
where $\mathrm{x}$ represents the generated images, $\mathrm{X}$ is the distribution of generated images.
$p(\mathrm{y}|\mathrm{x})$ is the conditional probability distribution of the class $\mathrm{y}$ given the image $\mathrm{x}$, predicted by the Inception model, $p(\mathrm{y})$ is the marginal probability of class $\mathrm{y}$.
$D_{KL}$ is the Kullback-Leibler divergence.
\item \textbf{Precision and Recall}~\cite{precrecall} are used to evaluate the quality of generated images by comparing them to real images. Precision measures how much the generated images resemble real images, while Recall measures how much of the real image distribution is captured by the generated images.
\item \textbf{Centered Kernel Alignment} (CKA) is a widely adopted metric for quantifying neural network representations~\cite{davari2022reliability, kornblith2019similarity}, which has been demonstrated with several advantages:
1) CKA is invariant to orthogonal transformation and isotropic scaling, thus it is stable under various image transformations;
2) CKA can capture the non-linear correspondence between representations benefit from its kernel mapping in the kernel space;
and 3) CKA can quantify the correspondence between different features across different widths, whereas previous metrics fail~\cite{kornblith2019similarity}.
Formally, CKA is normalized from Hilbert-Schmidt Independence Criterion (HSIC)~\cite{HSIC} to be invariant to orthogonal
transformation and isotropic scaling:
\begin{align}
    \label{eq:cka}
    \mathrm{CKA(X,Y)}=\frac{\mathrm{HSIC}(\mathrm{x},\mathrm{y})}{\sqrt{\mathrm{HSIC}(\mathrm{x},\mathrm{x}) \mathrm{HSIC}(\mathrm{y},\mathrm{y})}}.
\end{align}
HSIC identifies whether two distributions ($\mathrm{X,Y}$) are independent: $\mathrm{HSIC}(K,L) \!=\! \frac{1}{(n-1)^2}\operatorname{Tr}(K H L H)$, where $K_{i j}\!=\!k\left(\mathrm{x}_i, \mathrm{x}_j\right)$ { and } $L_{i j}\!=\!l\left(\mathrm{y}_i, \mathrm{y}_j\right)$, where $k$ and $l$ are kernels.
Note that for kernel selections of $k$ and $l$ in \cref{eq:cka}, we find that different kernels (RBF, polynomial, and linear) reflect similar discrepancies across various representations of DiTs, while the RBF kernel contributes to the distinguishability of quantitative results.
\end{itemize}

\begin{table*}[t]
    \centering
    \small
    \vskip -0.1in
    \caption{\textbf{Detailed quantitative results of our systematic analysis}. 
    All implementation details strictly follow the default settings of SiT and REPA for our investigation.
    All baselines are reported using vanilla-REPA \cite{repa} for training.
    }
    \vskip -0.1in
    \setlength{\tabcolsep}{2mm}{
    \begin{tabular}{lccccccccc}
        \toprule
        \textbf{Model} & \textbf{Alignment?}  & \textbf{Encoder}. & \textbf{Align Depth}. &\textbf{Iter}. & \textbf{FID}$_\downarrow$ & \textbf{sFID}$_\downarrow$ & \textbf{IS}$_\uparrow$ & \textbf{Prec.}$_\uparrow$ & \textbf{Rec.}$_\uparrow$ \\
        \midrule
        \textbf{SiT-B} \\
           & \xmark & \xmark  & \xmark & 50k  & 89.65   & 12.58  & 18.28   & 0.34    & 0.43 \\
           & \xmark & \xmark  & \xmark & 100k & 40.46   & 6.15   & 36.26   & 0.52    & 0.49 \\
           & \xmark & \xmark  & \xmark & 200k & 31.42   & 5.87   & 58.08   & 0.61    & 0.53 \\
           & \xmark & \xmark  & \xmark & 400k & 12.69   & 5.29   & 106.21  & 0.71    & 0.54 \\
           & \xmark & \xmark  & \xmark & 450k & 12.04   & 5.24  & 110.19   & 0.71    & 0.54 \\
        \midrule
        \textbf{REPA} \\
           & \cmark & DINOv2-B       & 5       & 450k & 5.37    & 5.35  & 175.07   & 0.75    & 0.58 \\
           & \cmark & DINOv2-B       & 8       & 450k & 7.67    & 5.60  & 150.87   & 0.72    & 0.58 \\
           & \cmark & DINOv2-B       & 10      & 450k & 10.85   & 6.12  & 128.34   & 0.70    & 0.58 \\
           & \cmark & DINOv2-B       & [2,5,8] & 450k & 13.25   & 5.27  & 105.64   & 0.70    & 0.55 \\
           & \cmark & DINOv2-B       & [3,6,9] & 450k & 13.54   & 5.50  & 104.88   & 0.69    & 0.56 \\
           & \cmark & MAE            & 5       & 450k & 10.15   & 5.11  & 123.24   & 0.72    & 0.55 \\
           & \cmark & MAE            & 8       & 450k & 11.40   & 5.22  & 115.20   & 0.72    & 0.55 \\
           & \cmark & MAE            & 10      & 450k & 12.11   & 5.30  & 111.01   & 0.71    & 0.55 \\
           & \cmark & DINOv2, MAE    & 5       & 450k & 5.77    & 5.10  & 166.15   & 0.76    & 0.57 \\
           & \cmark & DINOv2, MAE    & 8       & 450k & 7.44    & 5.37  & 150.12   & 0.73    & 0.57 \\
           & \cmark & DINOv2, MAE    & 10      & 450k & 11.03   & 6.17  & 124.89   & 0.70    & 0.55 \\
           & \cmark & DINOv2, MAE    & [3,8]   & 450k & 11.46   & 5.20  & 113.95   & 0.72    & 0.55 \\
           & \cmark & DINOv2, MAE    & [5,10]  & 450k & 11.73   & 5.22  & 111.77   & 0.71    & 0.54 \\
           & \cmark & DINOv2, MoCoV3 & [3,8]   & 450k & 11.36   & 5.24  & 115.75   & 0.71    & 0.55 \\
           & \cmark & DINOv2, MoCoV3 & [5,10]  & 450k & 12.71   & 6.08  & 104.46   & 0.66    & 0.56 \\
        \bottomrule
    \end{tabular}
    }
    \label{tab:supp_analysis_results}
\end{table*}

\section{Comparison Baselines}
\label{sec:supp:comparison_baselines}

In this part, we briefly introduce the main concept of baseline methods that are used for our evaluation.

\subsection{Multi-step baseline models}
\begin{itemize}[leftmargin=0.2in]
\item \textbf{ADM}~\cite{dhariwal2021diffusion} achieved improved synthesis performance with architectural improvement on traditional Unet-based diffusion models and developed classifier guidance to improve the synthesis fidelity for class-conditional tasks.
\item \textbf{VDM++}~\cite{kingma2023understanding} demonstrated that commonly used diffusion model objectives equate to a weighted integral of ELBOs over different noise levels, where the weighting depends on the specific objective used. Based on this, a sample adaptive noise schedule was introduced for improved training efficiency.
\item \textbf{CDM}~\cite{ho2022cascaded} proposed a cascaded architecture that trains multiple models across different resolutions, starting from the lowest resolution to higher resolution.
\item \textbf{LDM}~\cite{rombach2022high} developed latent diffusion models that train diffusion in a low-dimensional compressed latent space to improve the training efficiency. Specifically, the images are first encoded into latent codes and then added noise for training, and the denoised latents are decoded back to pixel space for sampling. 
\item \textbf{MDTv2}~\cite{gao2023mdtv2} introduced an asymmetric encoder-decoder paradigm for efficient training of diffusion transformer. To stabilize the training and improve model performance, they further employ U-Net-like long-shortcuts in the encoder and dense input-shortcuts in the decoder.
\item \textbf{MaskDiT}~\cite{zheng2023fast} used a similar encoder-decoder architecture with MDTv2, while the model was trained with an auxiliary reconstruction objective like \cite{he2022masked} to reconstruct masked inputs.
\item \textbf{SD-DiT}~\cite{zhu2024sd} extended the reconstruction-based MaskDiT architecture, while introducing a self-supervised discrimination objective with a momentum encoder for improved training.
\item \textbf{DiT}~\cite{dit} proposed to replace the conventional Unet-based architectures with transformers and further explored different condition injection mechanisms for conditional generation.
\item \textbf{SiT}~\cite{sit} systematically investigated the connections between discrete diffusion to continuous flow matching and developed practical training configurations for achieving strong synthesis performance.
\item \textbf{REPA}~\cite{repa} connected diffusion training dynamics and representation learning, revealing that pretrained external guidance could facilitate the representation learning of diffusion transformers.
\item \textbf{REG}~\cite{wu2025representation} further advanced REPA with a decoupled representation alignment technique, which entangled image latents and class tokens to imporve the conditional discrimination capability.
\item \textbf{E2E-REPA}~\cite{leng2025repa} unlocked a end-to-end training paradigm for joint tuning both the VAE and diffusion models throughout the training process, improving the VAE itself and downstream generation performance simultaneously.
\item \textbf{SRA}~\cite{jiang2025no} leveraged representations from later layers with lower noise of the EMA teacher to guide representations of earlier layers with higher noise, enabling a scheme of self-alignment.
\item \textbf{DispLoss}~\cite{jiang2025no} introduced a regularized dispersive loss to encourage internal features to spread out in the embedding space, thus facilitating the model to learn informative representations.
\end{itemize}

\subsection{One-step baseline models}
\begin{itemize}[leftmargin=0.2in]
\item \textbf{MeanFlow}~\cite{geng2025mean} introduced average velocity that was defined as the ratio of displacement to a time interval, with displacement given by the time integral of the instantaneous velocity. An intrinsic relation between the average and instantaneous velocities was then derived to guide efficient and effective one-step generative training.
\item \textbf{Shortcut}~\cite{shortcut} enhanced the few-step flow matching by adding a self-consistency loss, designed to learn the relationships between flow behaviors observed at different discrete time points.
\item \textbf{IMM}~\cite{inductive} learned a model that enforces self-consistency among stochastic interpolants evaluated at different points in time.
\item \textbf{iCT}~\cite{ict} leveraged consistency constraints across network outputs at different time steps to ensure that they predict the same endpoints along the trajectory.
\end{itemize}

\section{Detailed Results of Our Analysis}
\label{sec:supp:analysis_details}

\paragraph{Detailed Quantitative Results}
\cref{tab:supp_analysis_results} presents the detailed quantitative results of our systematic analysis in~\cref{sec:revisting_repa}.
These quantitative results consistently reflect the findings in our analysis:
1) aligning external representations on more blocks (\emph{e.g.,} aligning DINOv2-B features on [2, 5, 6]-th blocks and [3, 6, 9]-th blocks) does not bring obvious performance improvements, indicating that indiscriminate alignment can be detrimental and reduce the overall diversity between blocks, such observation is also reflected by the CKA similarity heatmaps in~\cref{fig:analysis}.
2) aligning with earlier blocks (e.g., Block 5) generally results in better performance than aligning with later blocks (e.g., Block 10), as evidenced by the lower FID scores, which is also identified in the original REPA.
3) combining different external encoders (DINOv2 and MAE) on different blocks does not consistently improve performance,  further indicating that the representation diversity across blocks is a crucial factor for high-quality synthesis.
Together, the quantitative results and CKA similarity heatmaps in~\cref{fig:analysis} consistently reveal that the key for representation learning is increasing the discrepancies of block representations.
Which provides explainable motivations for our proposed method in explicitly encouraging the representation diversity from the perspective of input and internal features' correlations.

\begin{figure*}[t]
    \vskip -0.2in
    \begin{center}  \centerline{\includegraphics[width=\linewidth]{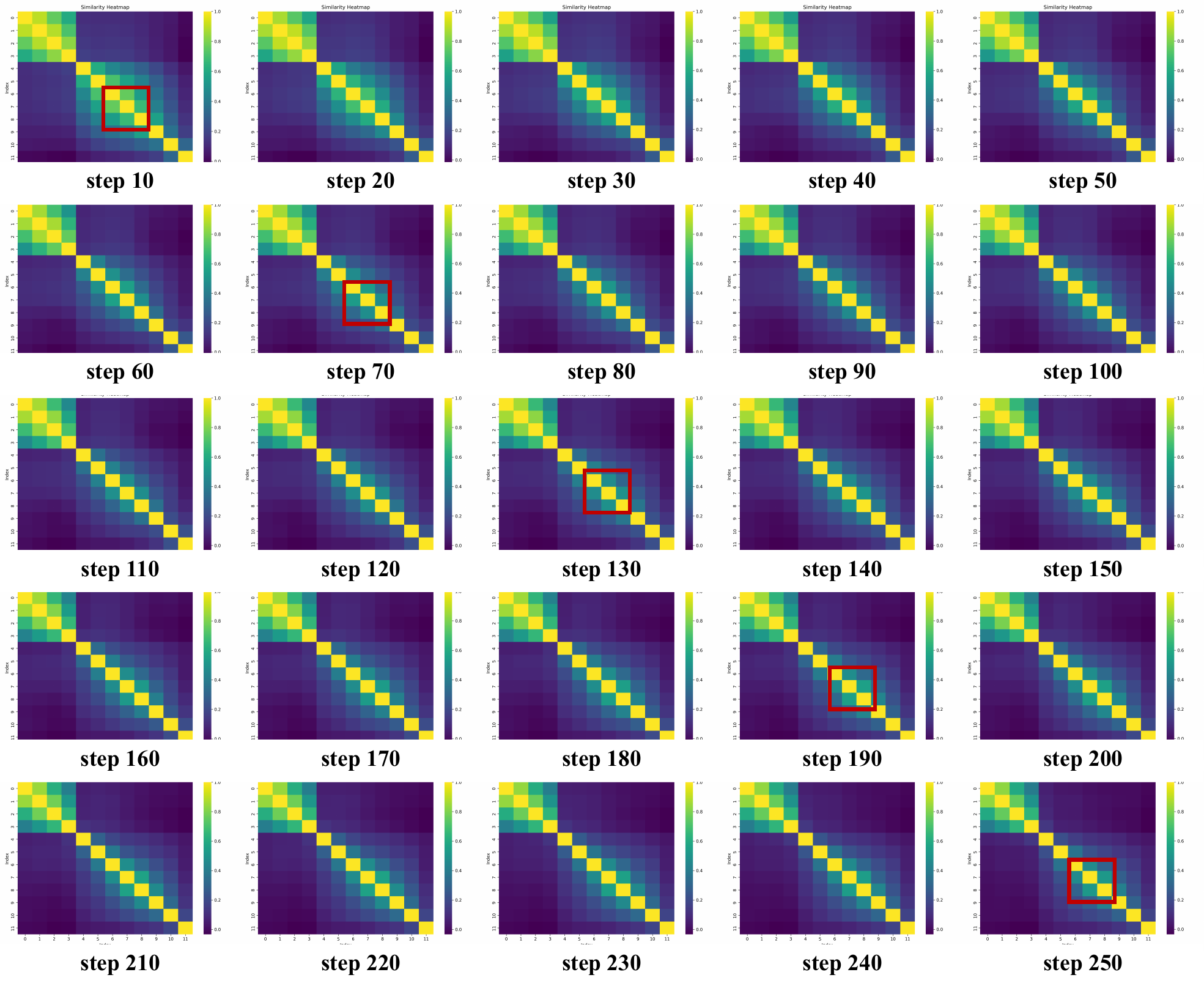}}
    \vskip -0.1in
    \caption{\textbf{CKA representation similarities across different timesteps.}  
    The representational discrepancies across different timesteps show similar correlations.
    }
    \label{fig:supp_cka_different_time}
    \end{center}
\end{figure*}

\paragraph{CKA similarity across various timesteps}
In~\cref{fig:analysis}, we present the CKA similarity heatmaps between representations of different blocks at the final denoising timestep.
To investigate the difference of block representations across different timesteps, we calculate their representational discrepancies of different timesteps in~\cref{fig:supp_cka_different_time}.
The results are computed from aligning DINOv2-B features on REPA-B for 400K training iterations like~\cref{fig:analysis}.
We could observe that the representation similarities between different blocks across different timesteps show a very similar pattern.
That is, the representational discrepancy across diffusion transformer blocks originates from the internal representation instead of different denoising timesteps.
Moreover, we can see that as the inference steps increases, the representational discrepancy between different blocks at different timesteps tends to slightly increase as well.
Such observation is reasonable because the noisy hidden states become less noisy throughout the sampling process.

\begin{table}[t]
    \centering
    \small
    \caption{\textbf{Variation in model-scale on ImageNet 512$\times$512 without CFG}. 
    Our proposed method brings consistent performance gains across all model-scales when applied to both SiT and REPA. 
    All baselines are reported using vanilla-REPA \cite{repa} for training.
    }
    \vskip -0.1in
    \setlength{\tabcolsep}{1.8mm}{
    \begin{tabular}{lcccccc}
        \toprule
        \textbf{Model}  & \textbf{Iter}. & \textbf{FID}$_\downarrow$ & \textbf{sFID}$_\downarrow$ & \textbf{IS}$_\uparrow$ & \textbf{Prec.}$_\uparrow$ & \textbf{Rec.}$_\uparrow$ \\
        \midrule
    SiT-B             & 400k  & 43.46          & 7.53          & 36.80          & 0.60          & 0.64           \\
    \textbf{+ (Ours)} & 400k  & \textbf{33.18} & \textbf{6.92} & \textbf{45.09} & \textbf{0.67} & \textbf{0.65}  \\
    \hdashline
    REPA-B            & 400k  & 30.13          & 7.79          & 53.92          & 0.68          & \textbf{0.64}  \\
    \textbf{+ (Ours)} & 400k  & \textbf{23.82} & \textbf{7.76} & \textbf{64.62} & \textbf{0.70} & 0.63           \\
    \midrule
    SiT-L             & 400k  & 22.75          & 5.78          & 64.05          & 0.73          & \textbf{0.63}  \\
    \textbf{+ (Ours)} & 400k  & \textbf{19.19} & \textbf{5.74} & \textbf{71.78} & \textbf{0.76} & 0.61           \\
    \hdashline
    REPA-L            & 400k  & 10.82          & 5.52          & 106.43         & 0.78          & 0.63           \\
    \textbf{+ (Ours)} & 400k  & \textbf{9.83}  & \textbf{5.49} & \textbf{114.00}& \textbf{0.78} & \textbf{0.64}  \\
    \midrule
    SiT-XL            & 400k  & 19.65          & 5.55          & 71.57          & 0.75          & \textbf{0.60}  \\
    \textbf{+ (Ours)} & 400k  & \textbf{17.68} & \textbf{5.48} & \textbf{76.45} & \textbf{0.77} & \textbf{0.60}  \\
    \hdashline
    REPA-XL           & 400k  & 7.91           & 5.41          & 127.83         & \textbf{0.79} & \textbf{0.65}  \\
    \textbf{+ (Ours)} & 400k  & \textbf{7.18}  & \textbf{5.38} & \textbf{137.09}& 0.78          & 0.64           \\
        \bottomrule
    \end{tabular}
    }
    \label{tab:supp_different_size_512}
\end{table}

\begin{table*}[t]
    \centering
    \small
    \caption{\textbf{Variation in alignment depth on ImageNet 256$\times$256 with different CFG scales}. 
    CFG=1.0 means no classifier-free guidance is applied.
    Our proposed method brings consistent performance gains across all model-scales when applied to REPA with different alignment depths and evaluated with different CFG scales.
    %
    }
    \vskip -0.1in
    \setlength{\tabcolsep}{1.8mm}{
    \begin{tabular}{lccccccccc}
        \toprule
        \textbf{Model} & \textbf{Iter}.  & \textbf{Encoder}. & \textbf{Align Depth}. &\textbf{CFG}. & \textbf{FID}$_\downarrow$ & \textbf{sFID}$_\downarrow$ & \textbf{IS}$_\uparrow$ & \textbf{Prec.}$_\uparrow$ & \textbf{Rec.}$_\uparrow$ \\
        \midrule
        \textbf{REPA-B} \\
           & 400K & DINOv2-B  & 5 & 1.0  & 22.99   & 6.70  & 64.73    & 0.59    & 0.65 \\
           & 400K & DINOv2-B  & 5 & 1.35 & 12.47   & 5.95  & 107.38   & 0.67    & 0.61 \\
         +\textbf{Ours} \\
           & 400K & DINOv2-B  & 5 & 1.0  & 17.29   & 6.56  & 79.92    & 0.62    & 0.65 \\
           & 400K & DINOv2-B  & 5 & 1.35 & 8.33    & 5.84  & 134.16   & 0.70    & 0.63 \\
        \hdashline
        \textbf{REPA-B} \\
           & 400K & DINOv2-B  & 8 & 1.0  & 27.94   & 7.19  & 54.32   & 0.56    & 0.64 \\
           & 400K & DINOv2-B  & 8 & 1.35 & 16.46   & 6.38  & 90.97   & 0.64    & 0.62 \\
         +\textbf{Ours} \\
           & 400K & DINOv2-B  & 8 & 1.0  & 23.27   & 6.82  & 62.63   & 0.59    & 0.65 \\
           & 400K & DINOv2-B  & 8 & 1.35 & 12.92   & 6.08  & 104.46  & 0.66    & 0.63 \\
        \midrule
        \textbf{REPA-L} \\
           & 400K & DINOv2-B  & 5 & 1.0  & 12.02   & 6.77  & 40.09   & 0.51    & 0.63 \\
           & 400K & DINOv2-B  & 5 & 1.35 & 5.14    & 4.74  & 157.71  & 0.75    & 0.61 \\
         +\textbf{Ours} \\ 
           & 400K & DINOv2-B  & 5 & 1.0  & 10.01   & 5.47  & 107.68   & 0.69    & 0.64 \\
           & 400K & DINOv2-B  & 5 & 1.35 & 4.18    & 4.61  & 172.26   & 0.76    & 0.61 \\
        \hdashline
        \textbf{REPA-L} \\
           & 400K & DINOv2-B  & 8 & 1.0  & 9.57    & 5.34  & 113.42   & 0.69    & 0.66 \\
           & 400K & DINOv2-B  & 8 & 1.35 & 3.86    & 4.82  & 183.18   & 0.75    & 0.63 \\
         +\textbf{Ours} \\
           & 400K & DINOv2-B  & 8 & 1.0  & 8.47    & 5.42  & 123.03   & 0.69    & 0.67 \\
           & 400K & DINOv2-B  & 8 & 1.35 & 3.39    & 4.80  & 196.08   & 0.76    & 0.63 \\
        \midrule
        \textbf{REPA-XL} \\
           & 400K & DINOv2-B  & 5 & 1.0  & 8.27    & 5.19  & 123.85   & 0.69    & 0.66 \\
           & 400K & DINOv2-B  & 5 & 1.35 & 3.33    & 4.73  & 196.52   & 0.75    & 0.64 \\
         +\textbf{Ours} \\
           & 400K & DINOv2-B  & 5 & 1.0  & 8.18    & 5.01  & 126.63   & 0.70    & 0.65 \\
           & 400K & DINOv2-B  & 5 & 1.35 & 3.17    & 4.71  & 198.30   & 0.77    & 0.62 \\
        \hdashline
        \textbf{REPA-XL} \\
           & 400K & DINOv2-B  & 8 & 1.0  & 8.73    & 5.21  & 118.68   & 0.69    & 0.65 \\
           & 400K & DINOv2-B  & 8 & 1.35 & 3.50    & 4.72  & 188.96   & 0.76    & 0.63 \\
         +\textbf{Ours} \\
           & 400K & DINOv2-B  & 8 & 1.0  & 8.09    & 5.02  & 123.23   & 0.70    & 0.65 \\
           & 400K & DINOv2-B  & 8 & 1.35 & 3.16    & 5.60  & 194.36   & 0.77    & 0.62 \\
        \bottomrule
    \end{tabular}
    }
    \vskip -0.15in
    \label{tab:supp_different_depth_cfg}
\end{table*}

\section{More Quantitative Results}
\label{sec:supp:more_quantitative}

\noindent \textbf{Improving representation learning across various model scales on ImageNet 512$\times$512.}
\cref{tab:supp_different_size_512} presents the quantitative results of applying our proposed techniques to SiT and REPA across various model scales on ImageNet 512$\times$512 without CFG.
Similar to the results of ImageNet 256$\times$256 in~\cref{tab:different_size}, our method consistently improves the performance of both SiT and REPA models across all scales, as evidenced by the reduction in FID and sFID scores and the increase in IS.
Specifically, when applied to SiT-B, our method achieves a significant improvement in FID score (from 43.46 to 33.18 and sFID from 7.53 to 6.92), while also improving the IS score from 36.80 to 45.09. 
Similar improvements can be observed for REPA-B, with FID improving from 30.13 to 23.82 and IS increasing from 53.92 to 64.62.
The benefits of our method are also evident for larger models. 
For SiT-XL, our approach reduces FID from 19.65 to 17.68 and increases IS from 71.57 to 76.45. For REPA-XL, the FID decreases from 7.91 to 7.18, and the IS increases from 127.83 to 137.09.
These results further indicate that our method is effective in improving the representation learning capabilities of both SiT and REPA models, regardless of their scale. 
The consistent improvements in FID, sFID, and IS across different model sizes demonstrate the robustness and generalizability of our approach. 
The improvements in Precision and Recall also suggest that our method leads to better alignment between the generated images and the real data distribution.

\begin{table}[t]
    \centering
    \small
    \vskip -0.1in
    \caption{\textbf{Variation in model-scale on ImageNet 512$\times$512 with CFG=1.35}. 
    Our proposed method brings consistent performance gains across all model-scales when applied to both SiT and REPA. 
    %
    }
    \vskip -0.1in
    \setlength{\tabcolsep}{1.8mm}{
    \begin{tabular}{lcccccc}
        \toprule
        \textbf{Model}  & \textbf{Iter}. & \textbf{FID}$_\downarrow$ & \textbf{sFID}$_\downarrow$ & \textbf{IS}$_\uparrow$ & \textbf{Prec.}$_\uparrow$ & \textbf{Rec.}$_\uparrow$ \\
        \midrule
    SiT-B             & 400k  & 32.77          & 6.95          & 50.85          & 0.67          & \textbf{0.62}           \\
    \textbf{+ (Ours)} & 400k  & \textbf{23.96} & \textbf{6.45} & \textbf{63.07} & \textbf{0.73} & 0.61  \\
    \hdashline
    REPA-B            & 400k  & 21.27          & 7.34          & 78.25          & 0.73          & 0.62  \\
    \textbf{+ (Ours)} & 400k  & \textbf{16.44} & \textbf{7.30} & \textbf{92.11} & \textbf{0.75} & \textbf{0.63}           \\
    \midrule
    SiT-L             & 400k  & 14.85          & \textbf{5.41}          & 91.49          & 0.78          & \textbf{0.60}  \\
    \textbf{+ (Ours)} & 400k  & \textbf{12.44} & \textbf{5.41} & \textbf{101.08} & \textbf{0.79} & 0.58           \\
    \hdashline
    REPA-L            & 400k  & 5.57          & \textbf{5.35}          & 158.39         & \textbf{0.80}          & 0.62           \\
    \textbf{+ (Ours)} & 400k  & \textbf{4.66}  & 5.58 & \textbf{173.94}& \textbf{0.80} & \textbf{0.64}  \\
    \midrule
    SiT-XL            & 400k  & 12.50          & 5.17          & 102.38          & 0.79          & \textbf{0.58}  \\
    \textbf{+ (Ours)} & 400k  & \textbf{11.24} & \textbf{5.28} & \textbf{107.73} & \textbf{0.81} & \textbf{0.58}  \\
    \hdashline
    REPA-XL           & 400k  & 4.30           & \textbf{5.09}          & 174.70         & \textbf{0.81} & 0.61  \\
    \textbf{+ (Ours)} & 400k  & \textbf{3.98}  & 5.48 & \textbf{184.73}& 0.80          & \textbf{0.62}           \\
        \bottomrule
    \end{tabular}
    }
    \vskip -0.15in
    \label{tab:supp_different_size_512_cfg}
\end{table}

\paragraph{Comparison results across different model scales with different CFG scales}
We mainly present comparison results without using classifier-free guidance (CFG)~\cite{nichol2021improved} in the main paper.
In this part, we present comparison results across different model scales with CFG enabled to further investigate its impact and our performance.
Specifically, we conduct experiments on ImageNet 256x256 on REPA using the DINOv2-B encoder for 400K training iterations across different model sizes (REPA-B, REPA-L, and REPA-XL).
We systematically evaluated the performance with different CFG scales (1.0, representing no classifier-free guidance, and 1.35).
\cref{tab:supp_different_depth_cfg} presents a detailed analysis of the impact of Classifier-Free Guidance (CFG) scale on the performance of our proposed method when applied to REPA models of varying sizes (REPA-B, REPA-L, and REPA-XL). 

First, the results consistently demonstrate that increasing the CFG scale from 1.0 to 1.35 leads to significant improvements in image quality and diversity across all model scales.
This is evidenced by the substantial increase in IS scores and the decrease in FID scores observed across all REPA model sizes when CFG is enabled.
Second, our proposed method also gains consistent performance improvement across different model scales when CFG is enabled.
For instance, our model advances the FID score of REPA-B from 12.47 to 8.33 and IS score from 107.38 to 134.16 with CFG=1.35, attaining a $>$32\% performance improvement on FID.
Similarly, our model advances the FID score of REPA-XL from 3.50 to 3.16 and the IS score from 188.96 to 194.36 with CFG=1.35.

Furthermore, \cref{tab:supp_different_size_512_cfg} presents the comparison results on on ImageNet 512$\times$512 with CFG=1.35.
Across all model scales, our method consistently improves the FID and sFID scores when CFG is used, indicating enhanced image quality and fidelity. 
For example, when applied to SiT-B, our method reduces the FID from 43.46 to 33.18 and the sFID from 7.53 to 6.92. Similarly, for REPA-B, the FID decreases from 30.13 to 23.82.
Together with the results that were tested without using CFG, these results demonstrate the scalability and effectiveness of our proposed method to higher resolutions and different model sizes.

\begin{table}[t]
    \centering
    \small
    \caption{\textbf{Ablation analysis on different components with CFG=1.35}.}
    \label{tab:ablation_components_supp}
    \vskip -0.1in
    \setlength{\tabcolsep}{2mm}{
    \begin{tabular}{lccccc}
        \toprule
         \textbf{Component} & \textbf{FID}$_\downarrow$ & \textbf{sFID}$_\downarrow$ & \textbf{IS}$_\uparrow$ & \textbf{Prec.}$_\uparrow$ & \textbf{Rec.}$_\uparrow$ \\
         \midrule
        SiT-B Baseline               & 23.28         & 6.00          & 65.23         & 0.61          & 0.60         \\
        SiT-B + ${\texttt{full}}$    & 16.21         & 5.45          & 84.00         & 0.66          & 0.60         \\
        w/o ${\texttt{diversity}}$   & 20.07         & 5.72          & 73.65         & 0.63          & 0.60         \\
        w/o ${\texttt{residual}}$    & 20.76         & 5.77          & 69.23         & 0.61          & 0.61         \\ \midrule
        REPA                         & 12.47         & 5.85          & 107.38        & 0.61          & 0.62         \\
        REPA-B + ${\texttt{full}}$   & 8.34          & 5.64          & 134.16        & 0.70          & 0.63        \\
        w/o ${\texttt{diversity}}$   & 10.75         & 5.75          & 115.42        & 0.68          & 0.62         \\
        w/o ${\texttt{residual}}$    & 11.02         & 5.77          & 112.93        & 0.64          & 0.62         \\
        \bottomrule
    \end{tabular}
    }
\end{table}

\paragraph{Quantitative results of applying our method on REPA with alignment on different blocks}
\cref{tab:supp_different_depth_cfg} also provides insight into the impact of the depth of alignment on the performance of our proposed method. 
We evaluated the models with alignment depths of 5 and 8, while keeping other parameters constant.
The results suggest that increasing the alignment depth from 5 to 8 can have varying effects depending on the model size, suggesting that the optimal alignment depth may depend on the interplay between model size.
Despite these variations, our method consistently improves upon the baseline REPA models when performing alignment on different blocks, with or without CFG.
For example, REPA-XL with our method and an alignment depth of 5 achieves an FID score of 3.17 at CFG 1.35, compared to 3.33 for the baseline. 
Similarly, the IS score improves from 196.52 to 198.30. 
This consistent trend of improvement, regardless of alignment depth, demonstrating the effectiveness of our approach in enhancing image generation. 
The consistent improvements observed across different alignment depths and model sizes further demonstrate the robustness and generalizability of our approach.

\section{More Ablation and Analysis Results}
\label{sec:supp:more_ablation}

\paragraph{Ablation on layer selection $\mathcal{P}$}
In our implementation for layer selection $\mathcal{P}$, we randomly select 10 layers to compute the diversity loss for experiments.
To investigate its impact, here we testify the impact of $\mathcal{P}$ on SiT-L (24 layers) for 400K training steps.
The results in Tab.~\ref{tab:ablation_layer_selection} show that selecting more layers improves the performance but increases the training time. 
In particular, selecting 10 layers yields a better trade-off between performance and efficiency.

\begin{table}[t]
    \centering
    \small
    \caption{\textbf{Ablation analysis on selecting different number of layers for diversity loss}}
    \label{tab:ablation_layer_selection}
    \vskip -0.1in
    \setlength{\tabcolsep}{1.8mm}
    {
    \begin{tabular}{l|cccccc}
\hline
\textbf{$\mathcal{P}$}  & \textbf{SiT-XL}  & \textbf{5}    & \textbf{10}   & \textbf{15}   & \textbf{20}   & \textbf{all} \\ \hline
FID$_\downarrow$  & 18.77  & 16.85 & 16.10 & 16.01 & 15.84 & \textbf{15.77} \\
IS$_\uparrow$     & 71.44  & 77.62 & 79.47 & 82.05 & 83.95 & \textbf{85.64} \\
Time (h)          & 18.66  & 19.90 & 21.02 & 23.45 & 25.96 & 28.50          \\
    \bottomrule
    \end{tabular}
    }
\end{table}

\paragraph{Ablation on different loss variants}
Here we further perform ablation on each loss component of the proposed diversity loss on SiT-B baseline.
The results in Tab.~\ref{tab:ablation_loss_sit} show the effectiveness of each loss, consistent with the findings of REPA results in Tab.~\ref{tab:ablation_loss}.
Specifically, using all components of the diversity loss (SiT-B + full) achieves the best performance and removing any single loss component degrades performance.

\begin{table}[t]
    \centering
    \small
    \caption{\textbf{Ablation analysis on different loss variants on SiT-B baseline.}}
    \label{tab:ablation_loss_sit}
    \vskip -0.1in
    \setlength{\tabcolsep}{2mm}{
    \begin{tabular}{lccccc}
        \toprule
         \textbf{Component} & \textbf{FID}$_\downarrow$ & \textbf{sFID}$_\downarrow$ & \textbf{IS}$_\uparrow$ & \textbf{Prec.}$_\uparrow$ & \textbf{Rec.}$_\uparrow$ \\
         \midrule
     SiT-B + ${\texttt{full}}$                        & 28.05   & 6.04  & 50.66  & 0.57  & 0.63 \\
     ${\texttt{only $\mathcal{L}_{\text{orth}}$}}$  & 31.32   & 6.45  & 47.09  & 0.56  & 0.63 \\
     ${\texttt{only $\mathcal{L}_{\text{MI}}$}}$    & 29.97   & 6.21  & 48.23  & 0.57  & 0.63 \\
     ${\texttt{only $\mathcal{L}_{\text{div}}$}}$   & 36.12   & 6.64  & 45.04  & 0.55  & 0.62 \\
        \bottomrule
    \end{tabular}
    }
\end{table}

\noindent \textbf{Ablation on designed components with CFG.}
\cref{tab:ablation_components_supp} presents the ablative results on the designed components of our DiverseDiT with CFG.
We could see that applying the CFG consistently improves the overall scores.
Similar to the results in \cref{tab:ablation_components}, the results clearly demonstrate the importance of both the representation diversity loss and the long residual connections for optimal performance.
Removing the diversity loss (w/o diversity) worsens the FID scores for both SiT-B (from 23.28 to 20.07) and REPA-B (from 12.47 to 10.75). 
Similarly, removing the long residual connections (w/o residual) also noticeably increases FID for both baseline models. 
Despite some performance degradation, we can observe that applying any of our proposed techniques to the baseline methods, \emph{i.e.,} REPA-B and SiT-B, brings substantial performance improvements.
For instance, with only long residual connections (w/o diversity), we achieve an FID of 20.07 on SiT-B and an FID of 10.75 on REPA-B, which are better than the original baseline results (23.28 for SiT-B and 12.47 for REPA-B).
Similar conclusions could be observed from the results of only diversity loss (w/o residual) as well.
These results confirm that both components of DiverseDiT play a crucial role in promoting diverse representation learning and improving the performance.

\noindent \textbf{Effect of diversity loss variant with CFG.}
\cref{tab:ablation_loss_supp} presents an ablation analysis on different loss variants with CFG=1.35. 
Similar to the previous results, the table demonstrates the importance of each loss component for optimal performance. Removing any of the loss components, namely $\mathcal{L}{orth}$, $\mathcal{L}{MI}$, or $\mathcal{L}{div}$, degrades the FID score compared to the REPA-B + full configuration (8.34). 
While using only $\mathcal{L}{orth}$ results in an FID of 10.98, using only $\mathcal{L}{MI}$ gives an FID of 10.78, and using only $\mathcal{L}{div}$ improves the FID to 8.59. 
These results confirm that each loss component plays a role in improving the model's performance, which is also reflected by the better results compared with the REPA baseline when each loss is used in isolation.

\noindent \textbf{Combining with existing methods for further improvement with CFG.}
\cref{tab:compatibility_supp} further explores the effect of combining our method with existing approaches, specifically DispLoss~\cite{wang2025diffuse} and SRA~\cite{jiang2025no}, on the SiT-B baseline with CFG=1.35.
Adding our method to the SiT-B baseline improves the FID from 23.28 to 16.21. 
Further combining with DispLoss results in an even lower FID of 13.73. 
This demonstrates that our method is complementary to existing techniques and can be combined with them to achieve further improvements in image generation quality.
Note that SRA and DispLoss require no additional external models for representation alignment, and combining our proposed method with them achieves a better performance than that of REPA, which needs pretrained models as guidance, demonstrating the potential for representation learning through internal mechanisms.

\begin{table}[t]
    \centering
    \small
    \caption{\textbf{Ablation analysis on different loss variants with CFG=1.35.}}
    \label{tab:ablation_loss_supp}
    \vskip -0.1in
    \setlength{\tabcolsep}{2mm}{
    \begin{tabular}{lccccc}
    \toprule
    \textbf{Component} & \textbf{FID}$_\downarrow$ & \textbf{sFID}$_\downarrow$ & \textbf{IS}$_\uparrow$ & \textbf{Prec.}$_\uparrow$ & \textbf{Rec.}$_\uparrow$ \\
    \midrule
REPA Baseline                                  & 12.47   & 5.85         & 107.38        & 0.61          & 0.62         \\
REPA-B + ${\texttt{full}}$                     & 8.34    & 5.64         & 134.16        & 0.70          & 0.63         \\
${\texttt{only $\mathcal{L}_{\text{orth}}$}}$  & 10.98   & 5.78         & 115.03        & 0.68          & 0.62         \\
${\texttt{only $\mathcal{L}_{\text{MI}}$}}$    & 10.78   & 5.76         & 115.95        & 0.69          & 0.63         \\
${\texttt{only $\mathcal{L}_{\text{div}}$}}$   & 8.59    & 5.77         & 131.69        & 0.70          & 0.63         \\
    \bottomrule
    \end{tabular}
    }
    \vskip -0.15in
\end{table}

\begin{table}[t]
    \centering
    \small
    \caption{
    \textbf{Combining our method with prior approaches with CFG=1.35.}
    }
    \label{tab:compatibility_supp}
    \vskip -0.1in
    \setlength{\tabcolsep}{1.8mm}{
    \begin{tabular}{lccccc}
        \toprule
         \textbf{Component} & \textbf{FID}$_\downarrow$ & \textbf{sFID}$_\downarrow$ & \textbf{IS}$_\uparrow$ & \textbf{Prec.}$_\uparrow$ & \textbf{Rec.}$_\uparrow$ \\
         \midrule
REPA Baseline                                  & 12.47   & 5.85         & 107.38        & 0.61          & 0.62         \\
\midrule 
SiT-B Baseline                                  & 23.28         & 6.00         & 65.23      & 0.61     & 0.60         \\
+ ${\texttt{Ours}}$                             & 16.21         & 5.45         & 84.00      & 0.66     & 0.60         \\
++ ${\texttt{DispLoss}}$~\cite{wang2025diffuse} & 13.73         & 5.76         & 95.31      & 0.68     & 0.60         \\
+++ ${\texttt{SRA}}$~\cite{jiang2025no}         & 11.25         & 5.37         & 108.15     & 0.69     & 0.61         \\
        \bottomrule
    \end{tabular}
    }
    \vskip -0.2in
\end{table}

\section{Limitations and Future Work}
\label{sec:supp:limitations}
%

\noindent \textbf{Limitations.}
Despite a comprehensive investigation, our analysis could be extended in several aspects:
For instance, whether \ourmethod can be effectively adapted to diverse generation tasks, such as text-to-image synthesis or image editing, remains an open question.
Besides, performing similar analysis on representation learning of other models like large-language models might reveal more interesting findings.
Additionally, we do not perform extensive hyperparameter searching for the optimal performance in our experiments, the full potential of our proposed representation diversity loss could be further unlocked.
Nevertheless, our study could provide potential guidelines for developing more effective methods in learning informative representations.

\noindent \textbf{Future work.}
For future work, we plan to extend our analysis and evaluation on text-to-image synthesis tasks.
We aim to investigate the application of our representation diversity loss to other generative models and modalities, such as video generation and 3D shape generation. 
Exploring different architectures and training strategies in conjunction with our proposed loss function could potentially lead to even more significant improvements in the quality and diversity of generated content. 
Meanwhile, we intend to explore theoretical connections between representation diversity and other desirable properties of generative models, such as robustness to adversarial attacks and generalization to unseen data distributions.
Furthermore, considering that our proposed diversity loss alone could likely be applied as a fine-tuning step for pre-trained models without any architectural changes, we plan to explore this in our ongoing research.

\section{More Qualitative Results}
\label{sec:supp:more_qualitative_results}
\begin{figure*}[t]
    \centering
    \includegraphics[width=\linewidth]{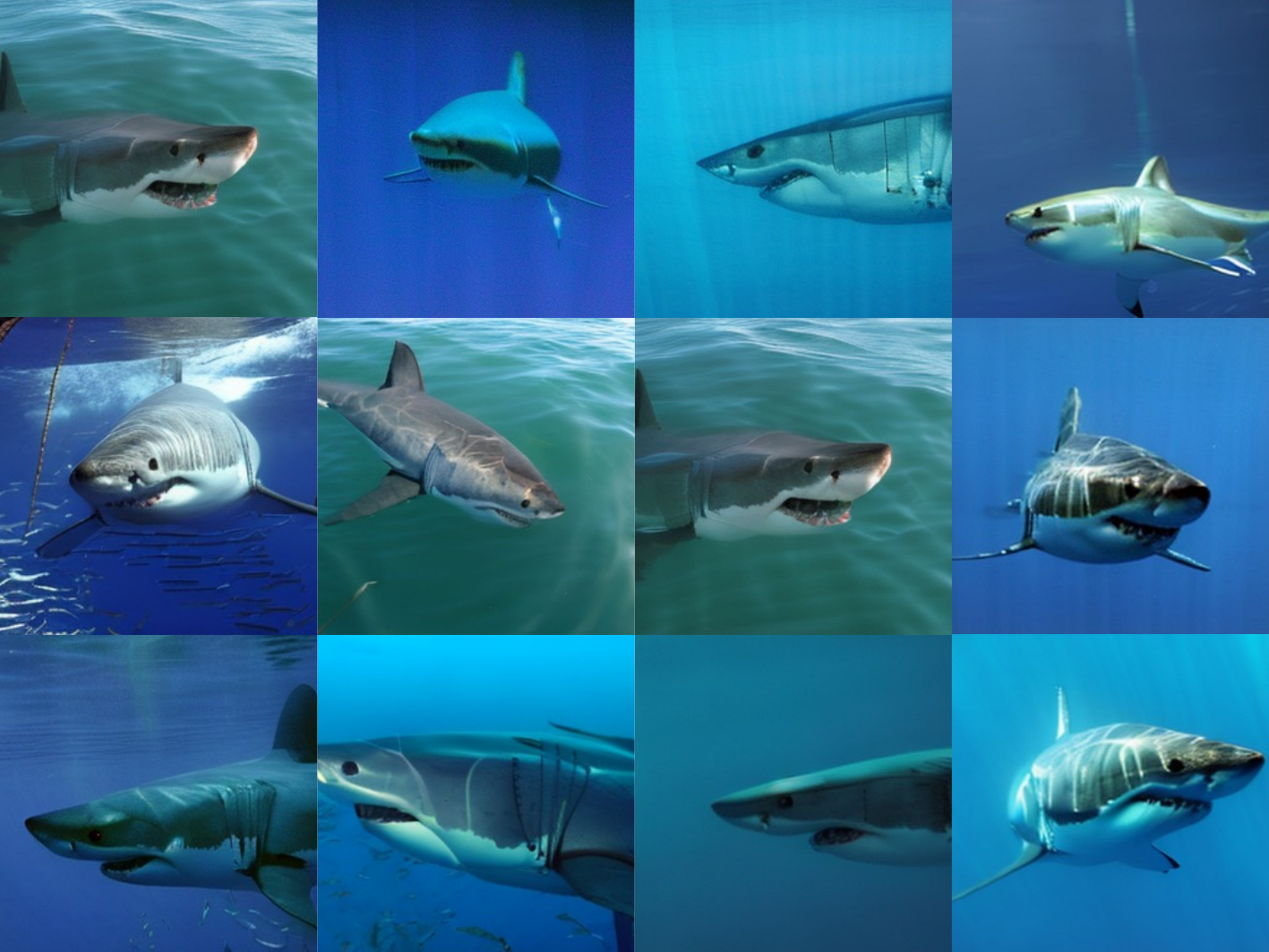}
    \caption{\textbf{Uncurated generation results of our DiverseDiT-XL on ImageNet 256$\times$256.} We use classifier-free guidance with $w=4.0$, the lass label is ``Great white shark'' (2).}
    \label{fig:supp_more_visualization_1}
\end{figure*}

\begin{figure*}[ht!]
    \centering
    \includegraphics[width=\linewidth]{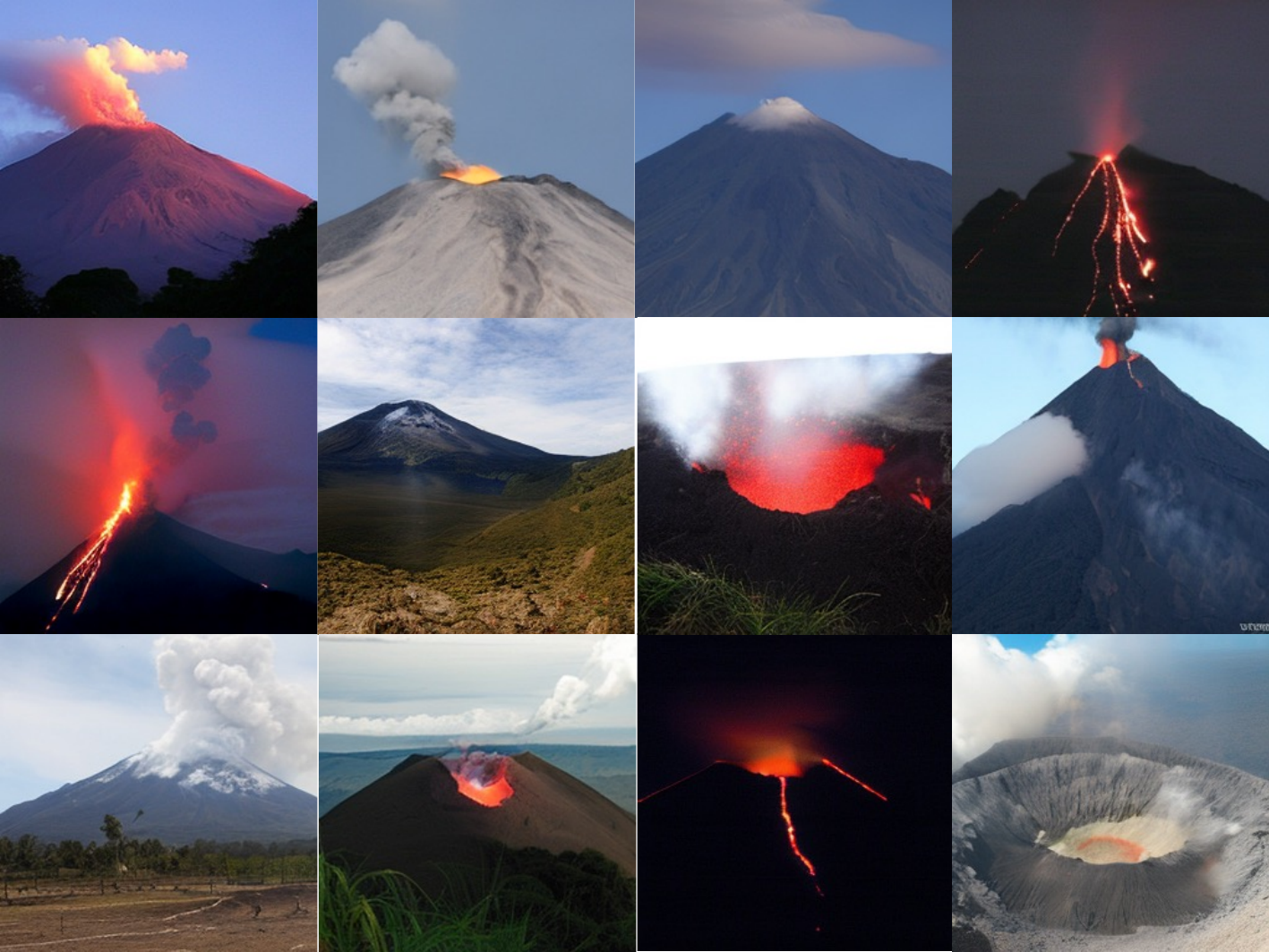}
    \caption{\textbf{Uncurated generation results of our DiverseDiT-XL on ImageNet 256$\times$256.} We use classifier-free guidance with $w=4.0$, the lass label is ``Volcano' (980).}
    \label{fig:supp_more_visualization_17}
\end{figure*}

We present more uncurated generation results of our DiverseDiT-XL on ImageNet 256$\times$256 in \cref{fig:supp_more_visualization_1} - \cref{fig:supp_more_visualization_17} with CFG (w = 4.0).

\begin{figure*}[ht!]
    \centering
    \includegraphics[width=\linewidth]{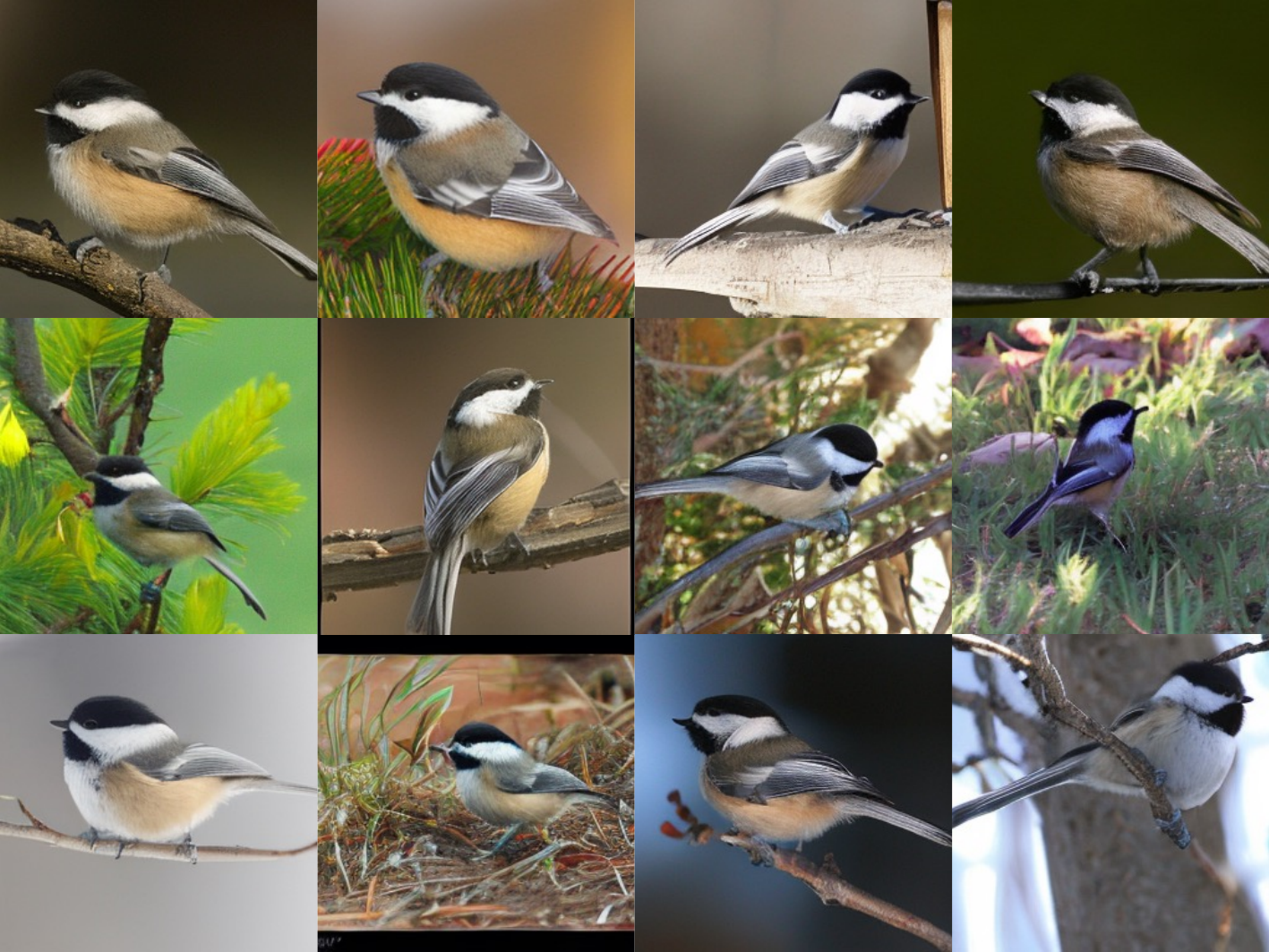}
    \caption{\textbf{Uncurated generation results of our DiverseDiT-XL on ImageNet 256$\times$256.} We use classifier-free guidance with $w=4.0$, the lass label is ``Chickadee'' (19).}
\end{figure*}
\begin{figure*}[ht!]
    \centering
    \includegraphics[width=\linewidth]{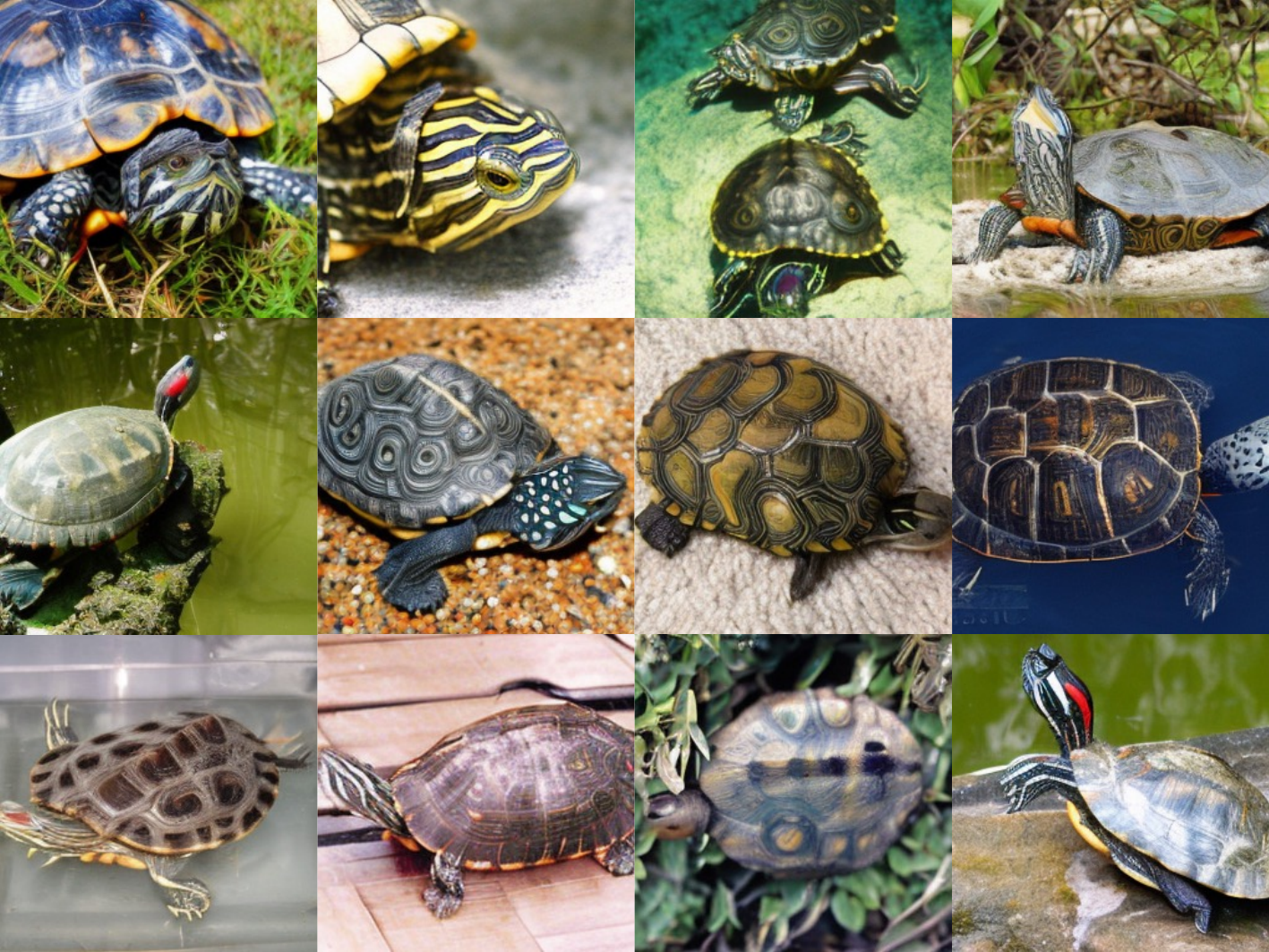}
    \caption{\textbf{Uncurated generation results of our DiverseDiT-XL on ImageNet 256$\times$256.} We use classifier-free guidance with $w=4.0$, the lass label is ``Terrapin'' (36).}
\end{figure*}
\begin{figure*}[ht!]
    \centering
    \includegraphics[width=\linewidth]{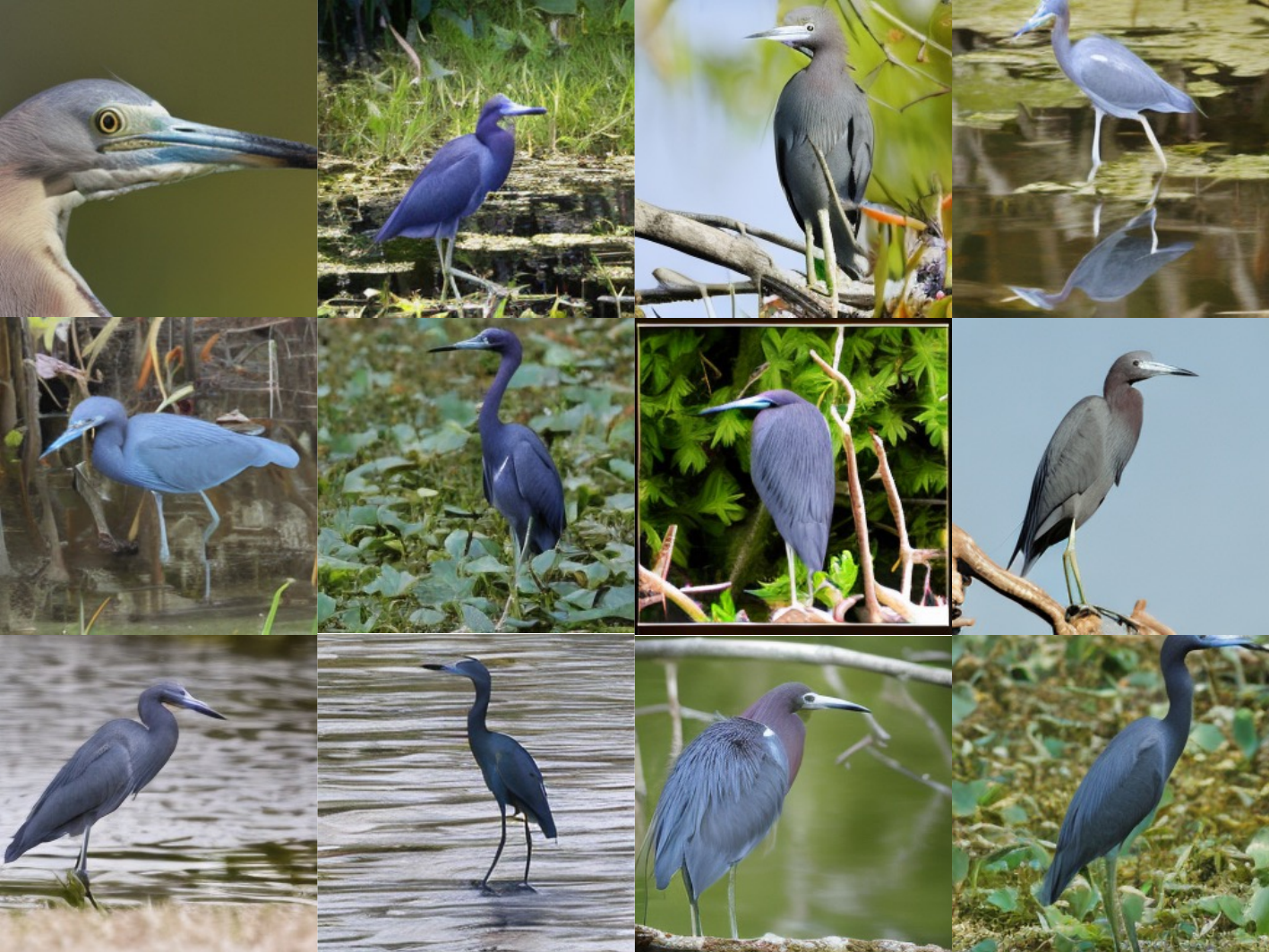}
    \caption{\textbf{Uncurated generation results of our DiverseDiT-XL on ImageNet 256$\times$256.} We use classifier-free guidance with $w=4.0$, the lass label is ``Little blue heron, Egretta caerulea'' (131).}
\end{figure*}
\begin{figure*}[ht!]
    \centering
    \includegraphics[width=\linewidth]{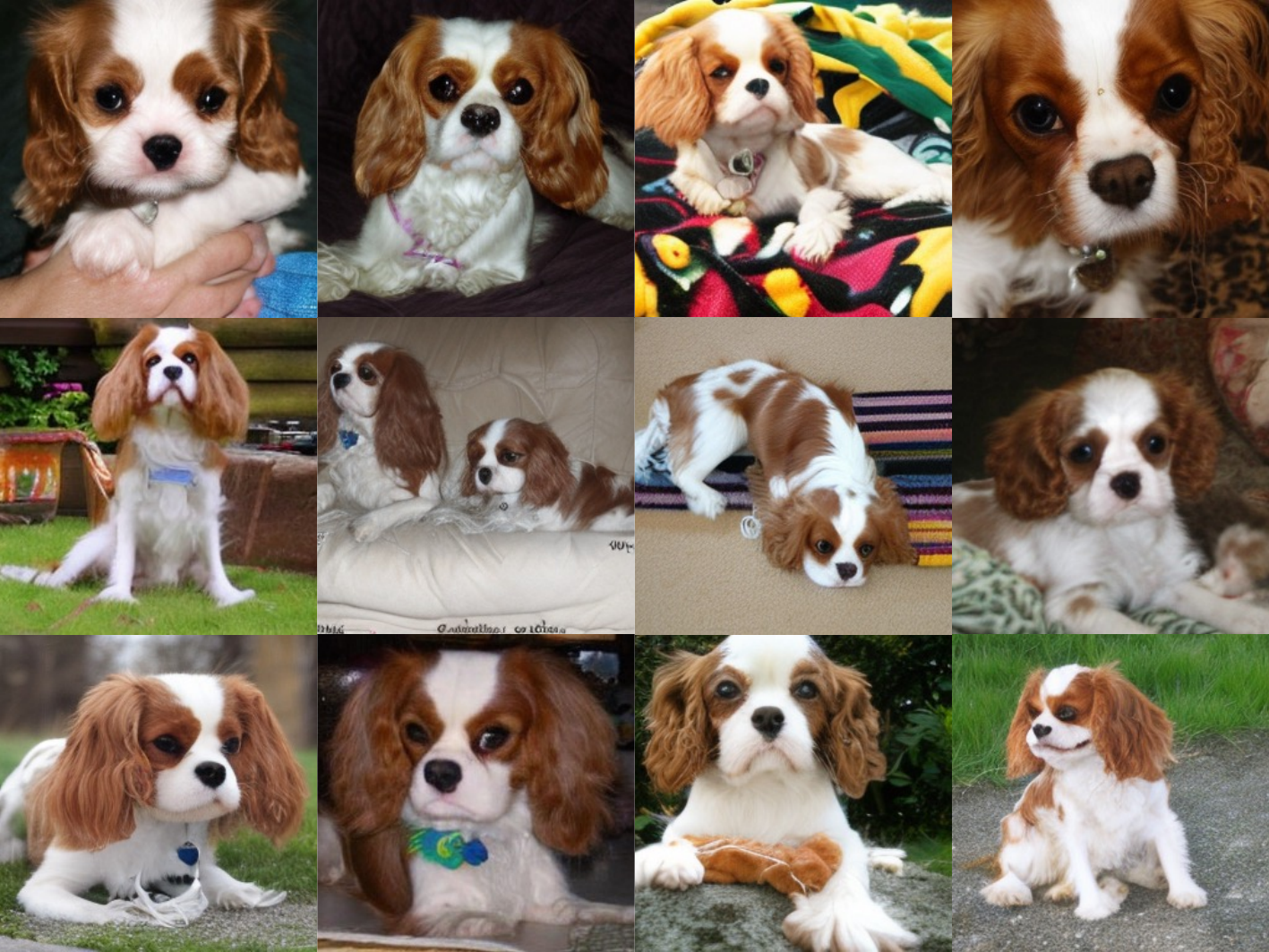}
    \caption{\textbf{Uncurated generation results of our DiverseDiT-XL on ImageNet 256$\times$256.} We use classifier-free guidance with $w=4.0$, the lass label is ``Blenheim spaniel'' (156).}
\end{figure*}
\begin{figure*}[ht!]
    \centering
    \includegraphics[width=\linewidth]{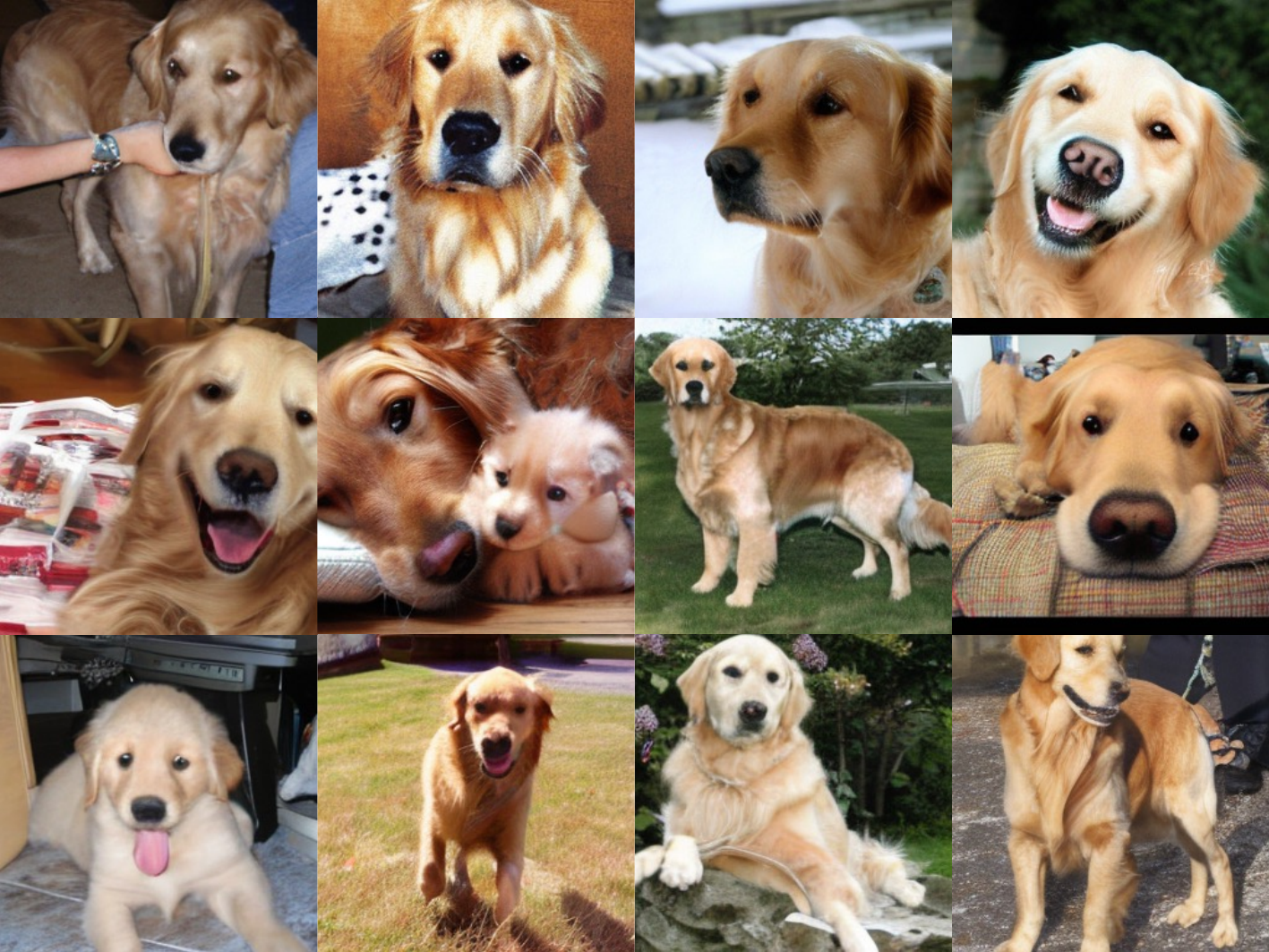}
    \caption{\textbf{Uncurated generation results of our DiverseDiT-XL on ImageNet 256$\times$256.} We use classifier-free guidance with $w=4.0$, the lass label is ``Golden retriever'' (207).}
\end{figure*}
\begin{figure*}[ht!]
    \centering
    \includegraphics[width=\linewidth]{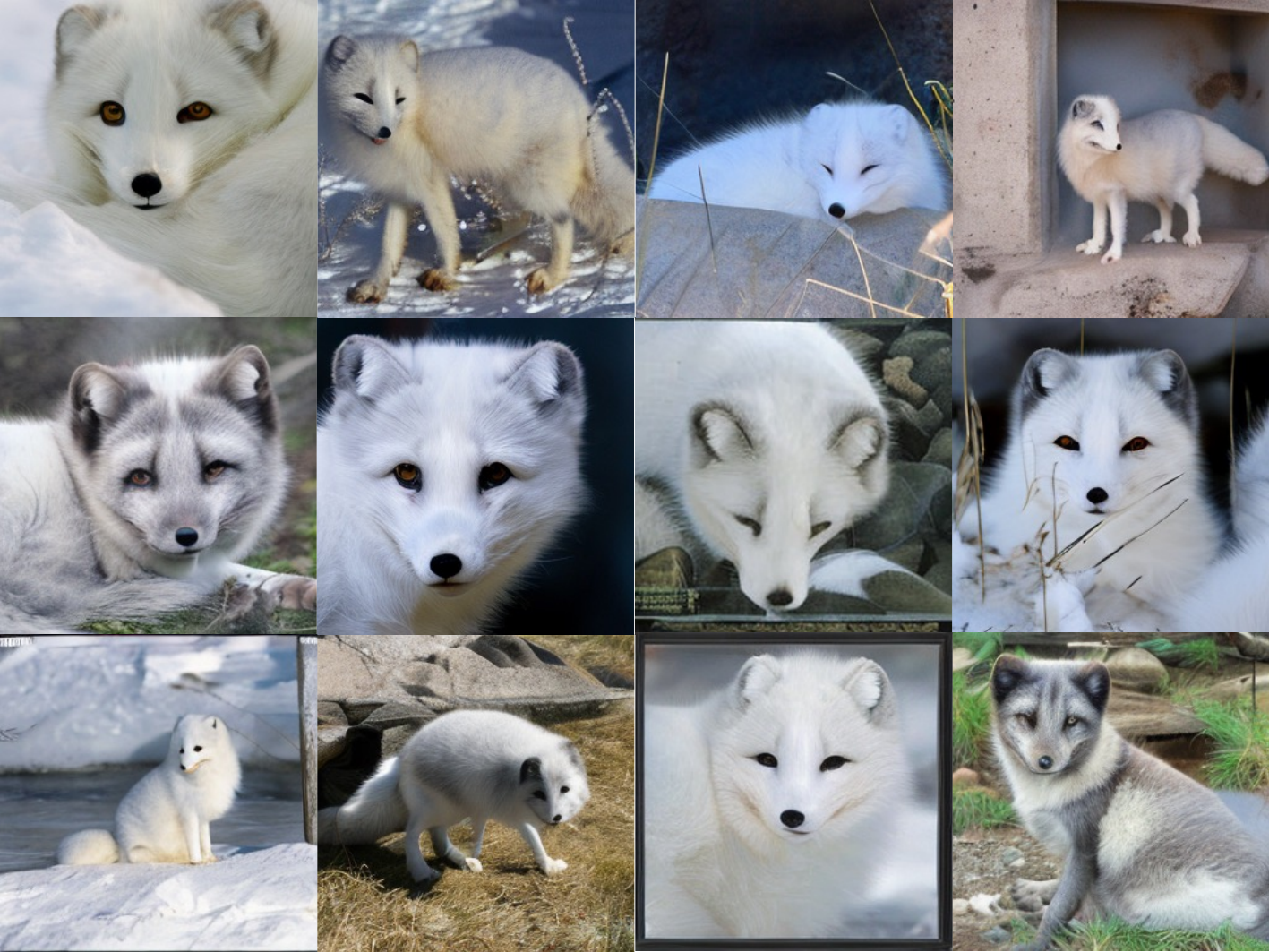}
    \caption{\textbf{Uncurated generation results of our DiverseDiT-XL on ImageNet 256$\times$256.} We use classifier-free guidance with $w=4.0$, the lass label is ``Arctic fox, White fox, Alopex lagopus'' (279).}
\end{figure*}
\begin{figure*}[ht!]
    \centering
    \includegraphics[width=\linewidth]{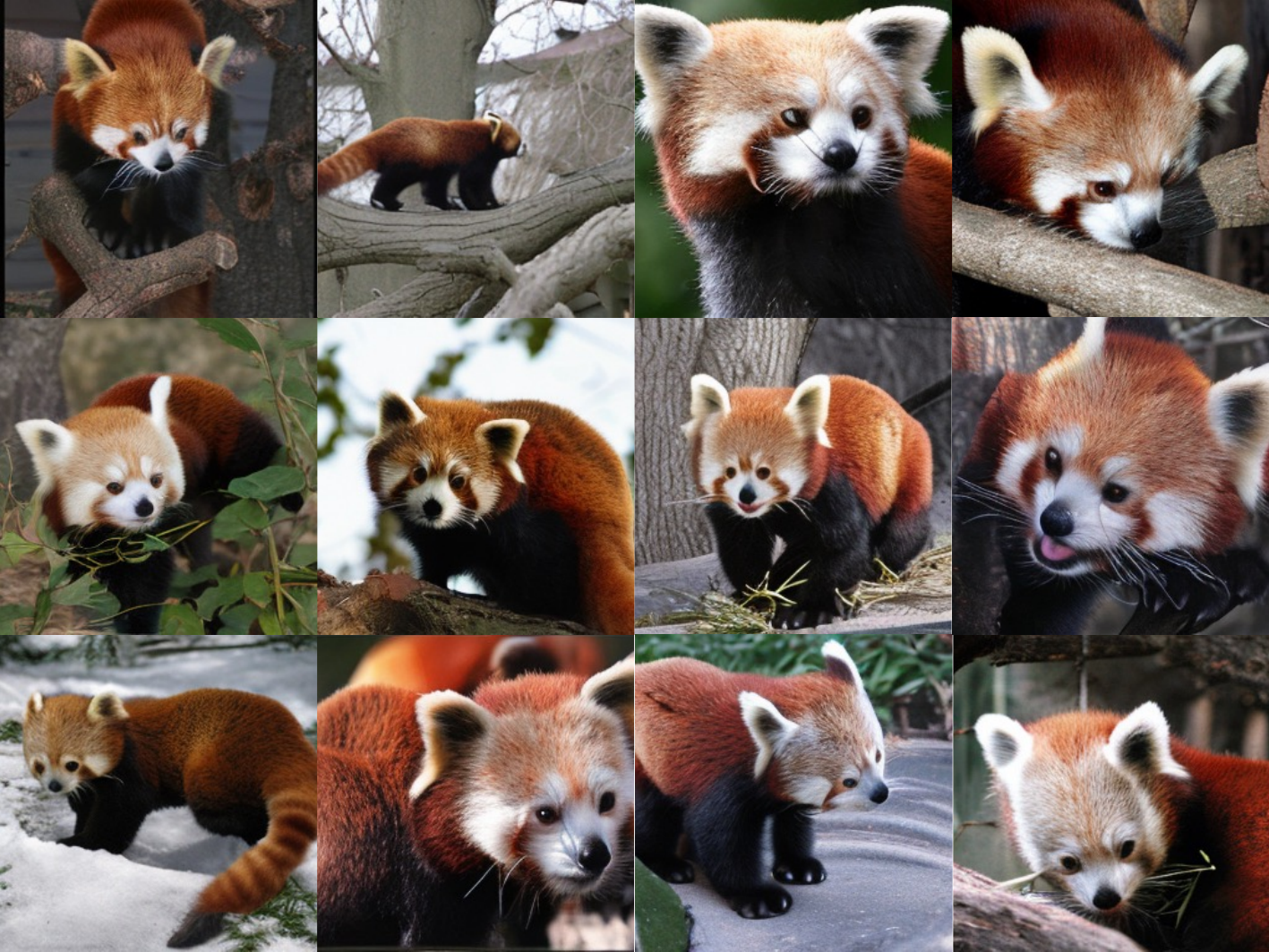}
    \caption{\textbf{Uncurated generation results of our DiverseDiT-XL on ImageNet 256$\times$256.} We use classifier-free guidance with $w=4.0$, the lass label is ``lesser panda, Red panda, Panda, Bear cat, Cat bear, Ailurus fulgens'' (387).}
\end{figure*}
\begin{figure*}[ht!]
    \centering
    \includegraphics[width=\linewidth]{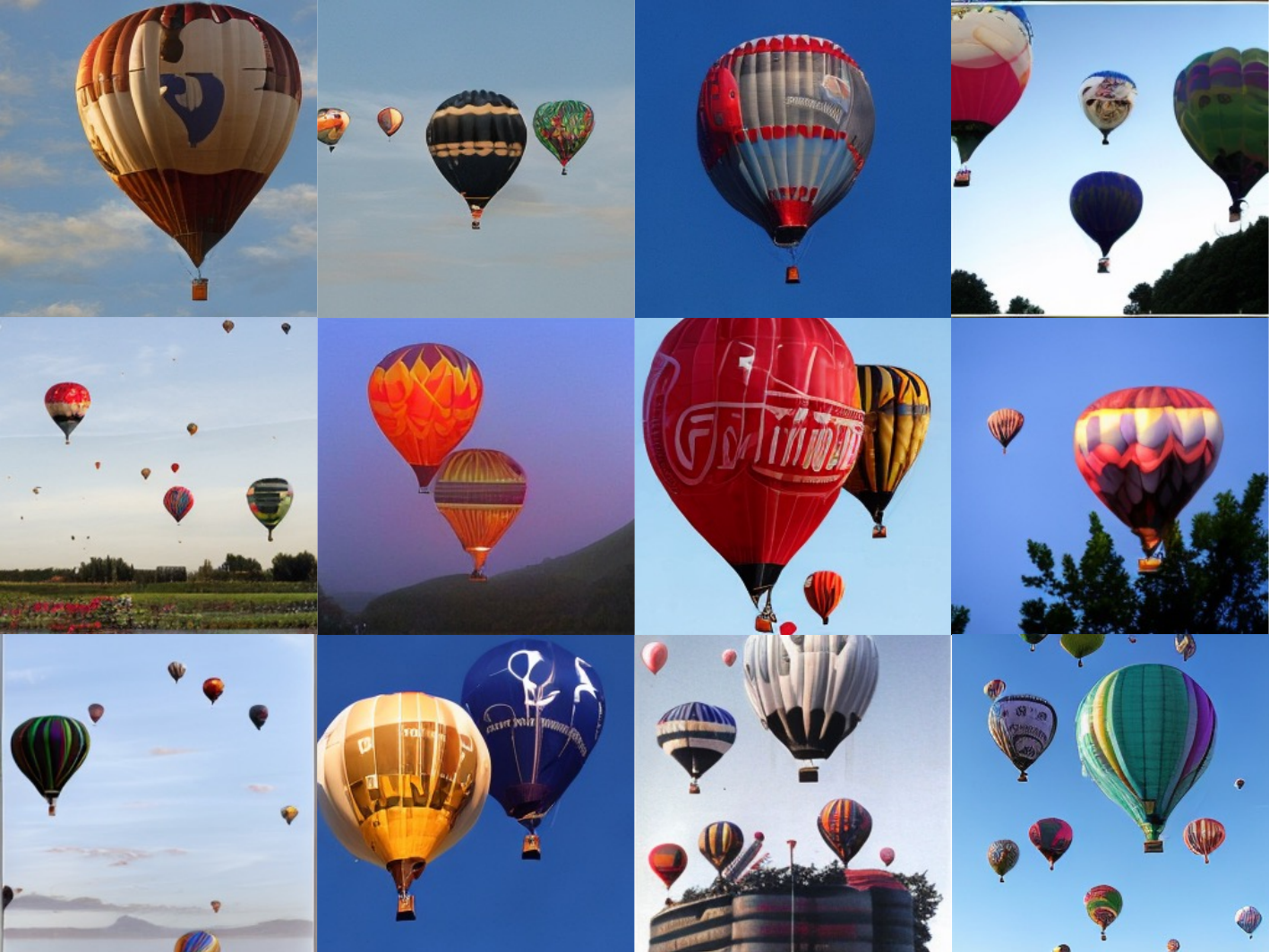}
    \caption{\textbf{Uncurated generation results of our DiverseDiT-XL on ImageNet 256$\times$256.} We use classifier-free guidance with $w=4.0$, the lass label is ``Balloon' (417).}
\end{figure*}
\begin{figure*}[ht!]
    \centering
    \includegraphics[width=\linewidth]{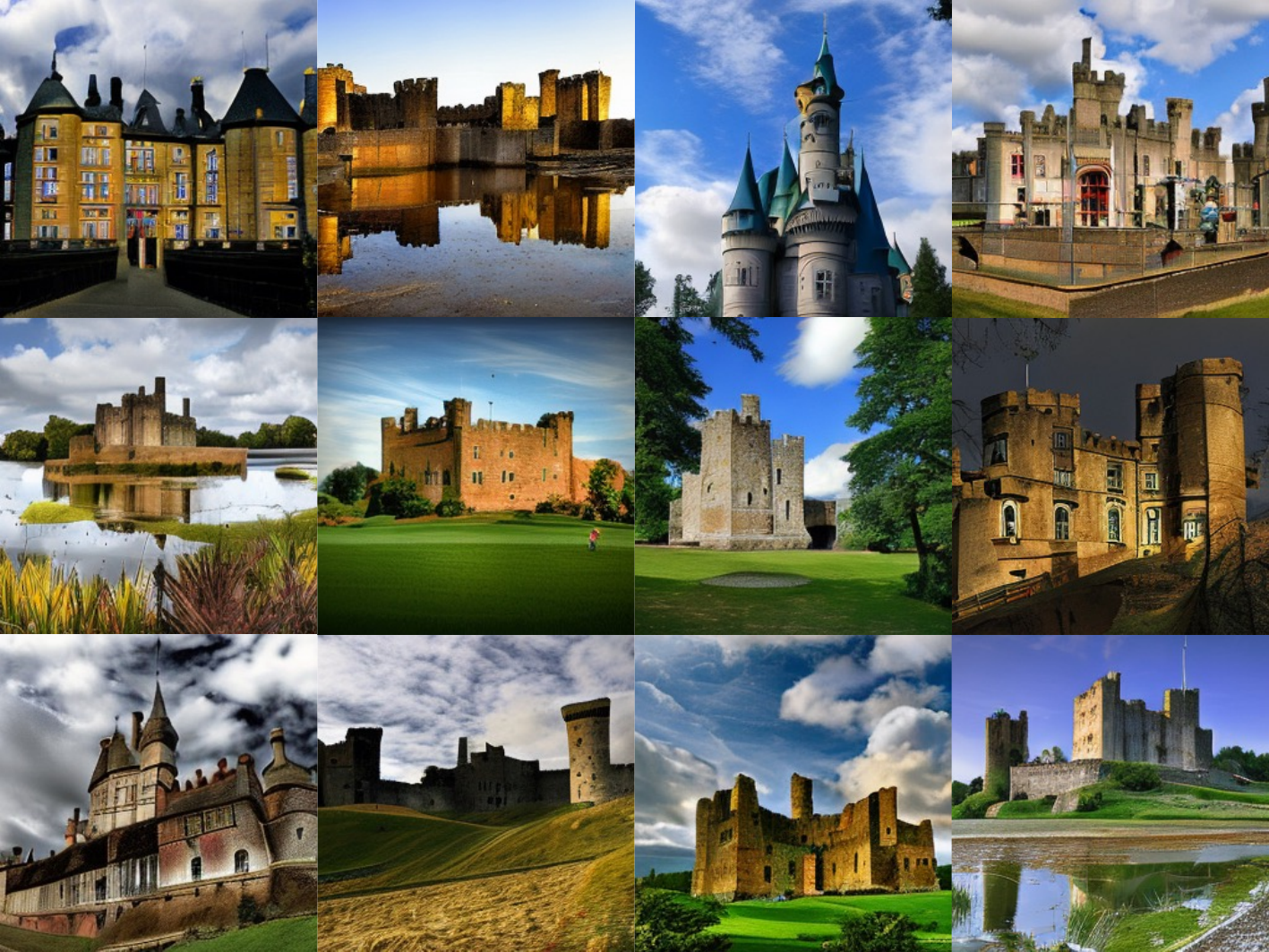}
    \caption{\textbf{Uncurated generation results of our DiverseDiT-XL on ImageNet 256$\times$256.} We use classifier-free guidance with $w=4.0$, the lass label is ``Castle' (483).}
\end{figure*}
\begin{figure*}[ht!]
    \centering
    \includegraphics[width=\linewidth]{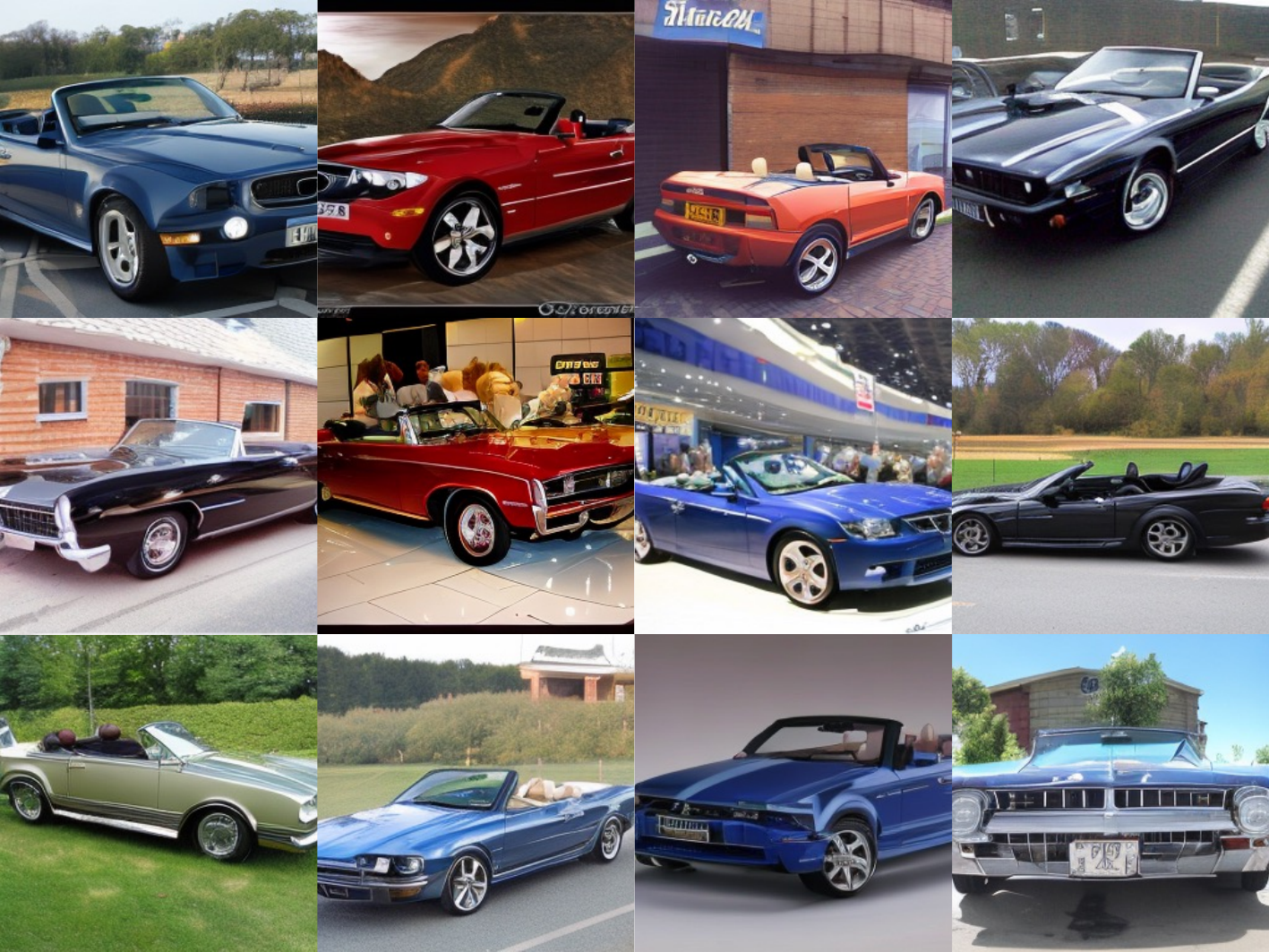}
    \caption{\textbf{Uncurated generation results of our DiverseDiT-XL on ImageNet 256$\times$256.} We use classifier-free guidance with $w=4.0$, the lass label is ``Check, Convertible' (511).}
\end{figure*}
\begin{figure*}[ht!]
    \centering
    \includegraphics[width=\linewidth]{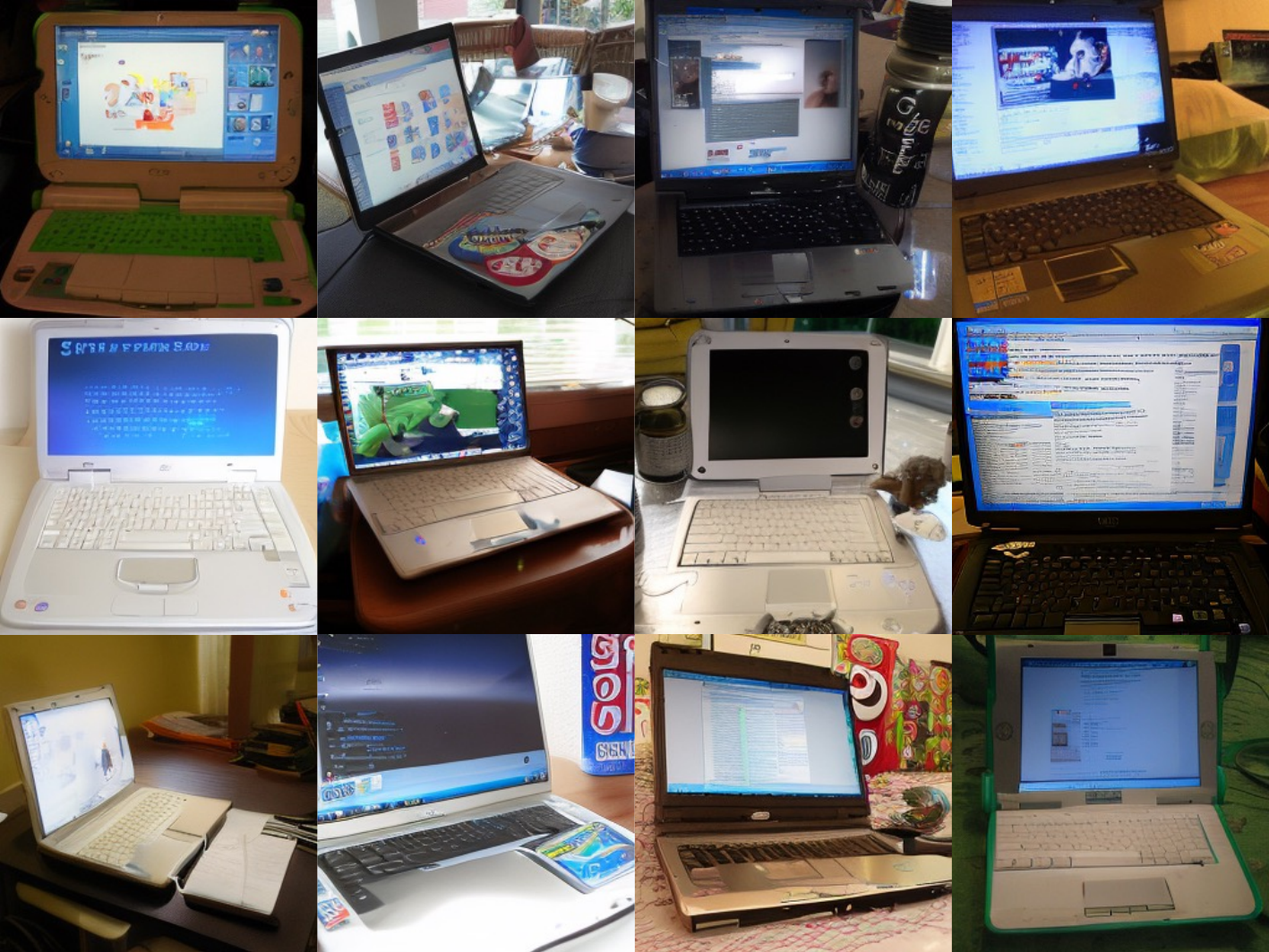}
    \caption{\textbf{Uncurated generation results of our DiverseDiT-XL on ImageNet 256$\times$256.} We use classifier-free guidance with $w=4.0$, the lass label is ``Laptop, Laptop computer' (620).}
\end{figure*}
\begin{figure*}[ht!]
    \centering
    \includegraphics[width=\linewidth]{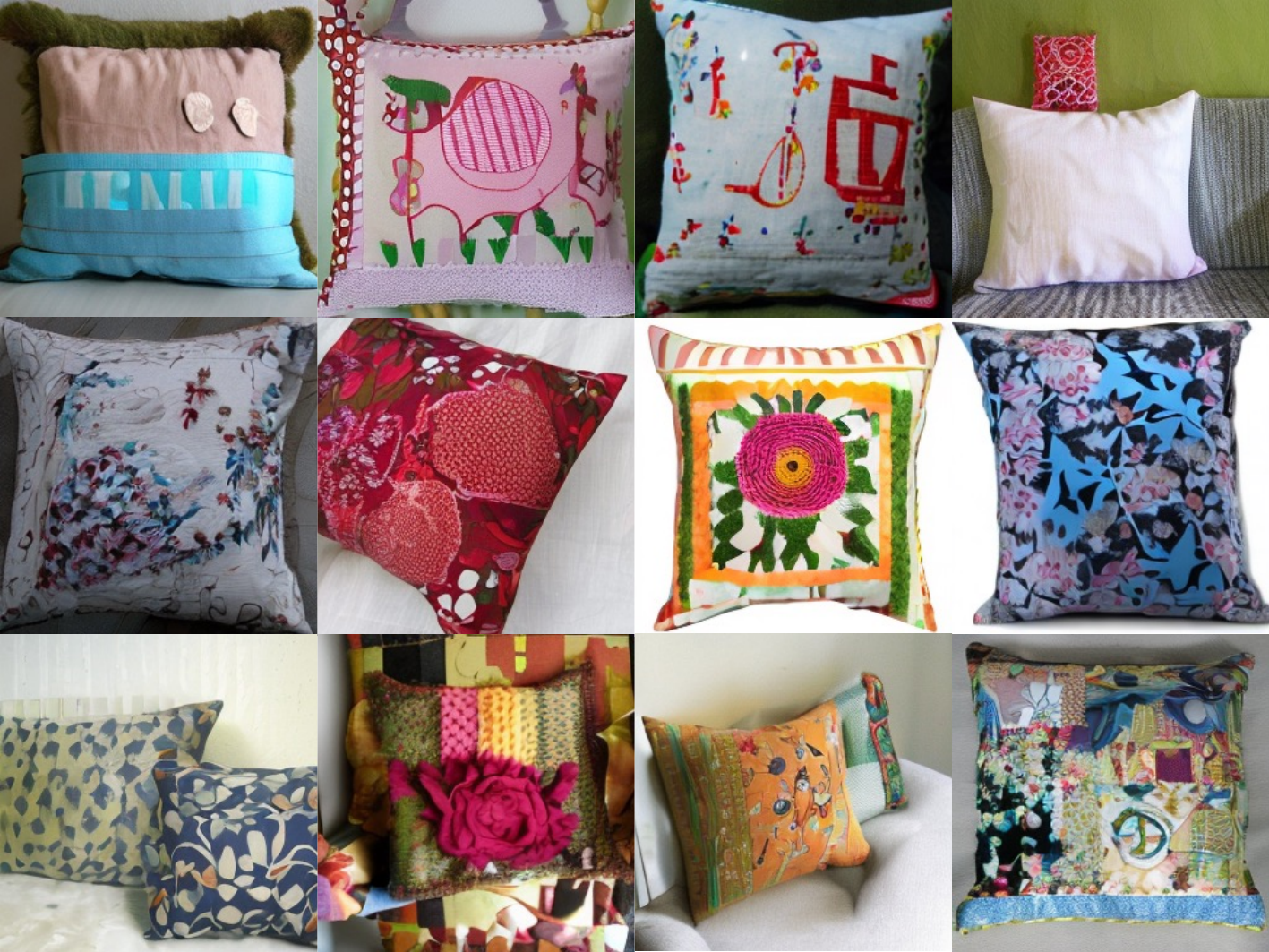}
    \caption{\textbf{Uncurated generation results of our DiverseDiT-XL on ImageNet 256$\times$256.} We use classifier-free guidance with $w=4.0$, the lass label is ``Pillow' (721).}
\end{figure*}
\begin{figure*}[ht!]
    \centering
    \includegraphics[width=\linewidth]{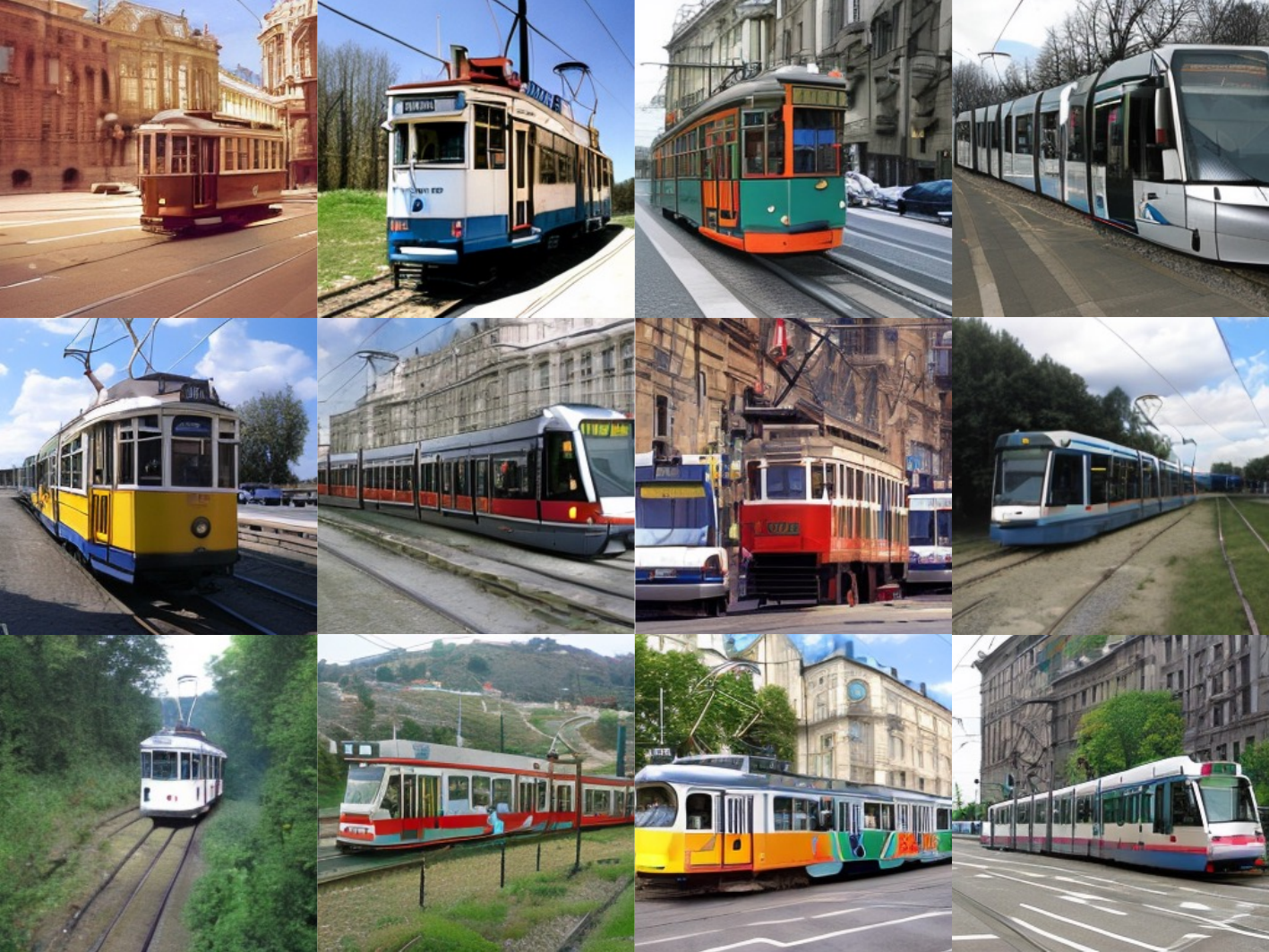}
    \caption{\textbf{Uncurated generation results of our DiverseDiT-XL on ImageNet 256$\times$256.} We use classifier-free guidance with $w=4.0$, the lass label is ``Check, Streetcar, Tram, Tramcar, Trolley, Trolley car' (829).}
\end{figure*}
\begin{figure*}[ht!]
    \centering
    \includegraphics[width=\linewidth]{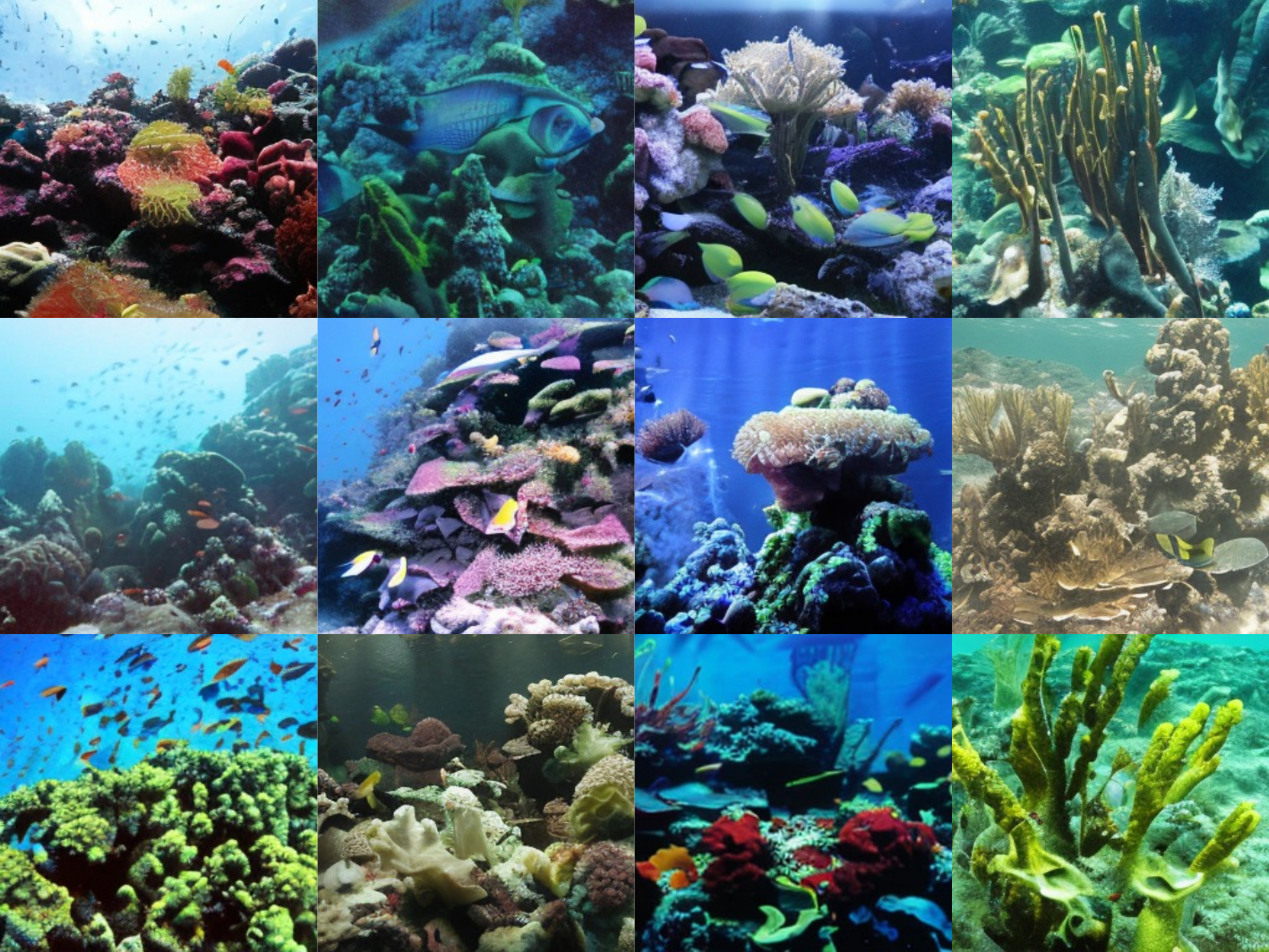}
    \caption{\textbf{Uncurated generation results of our DiverseDiT-XL on ImageNet 256$\times$256.} We use classifier-free guidance with $w=4.0$, the lass label is ``Coral reef' (973).}
\end{figure*}
\begin{figure*}[ht!]
    \centering
    \includegraphics[width=\linewidth]{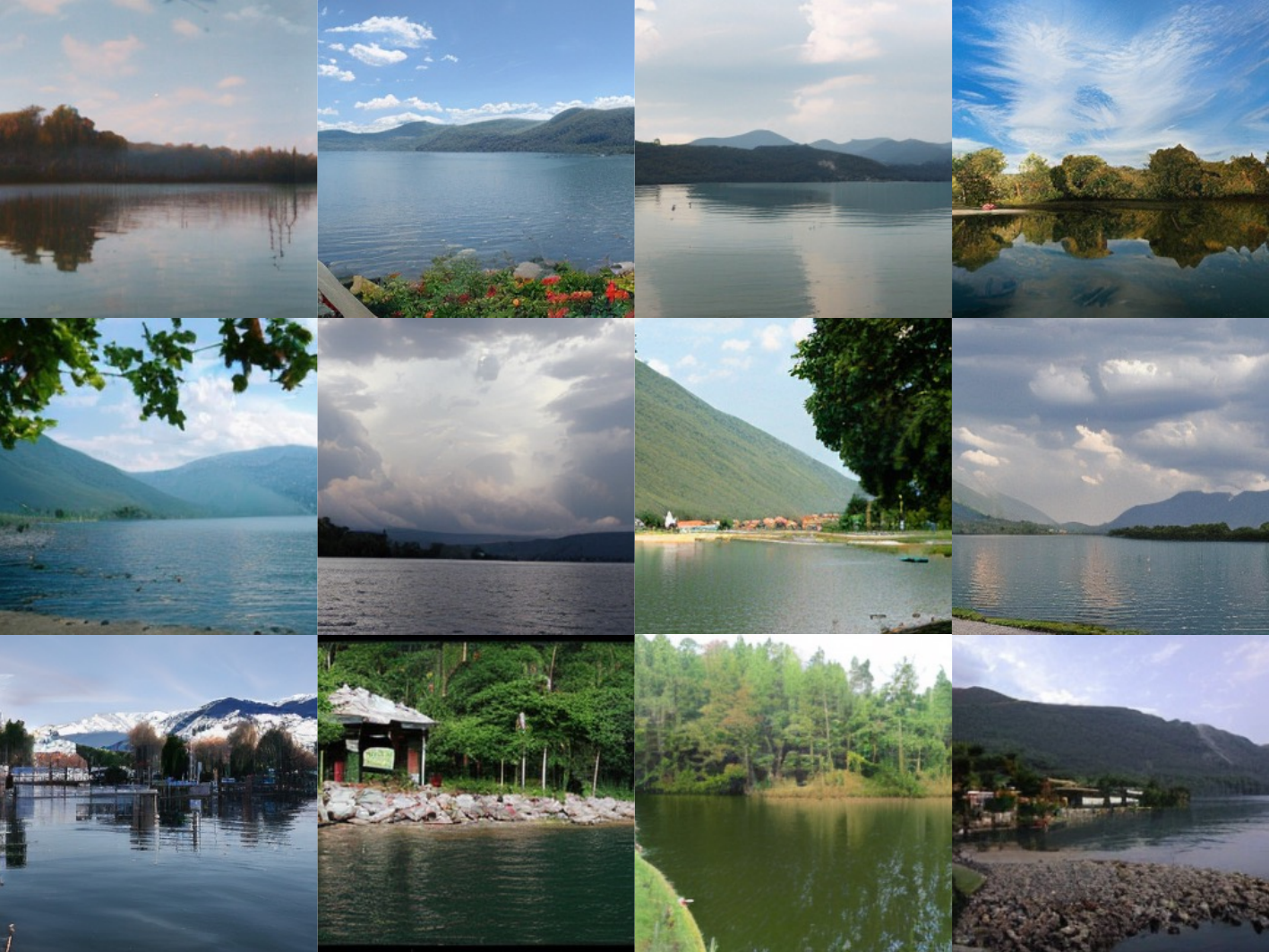}
    \caption{\textbf{Uncurated generation results of our DiverseDiT-XL on ImageNet 256$\times$256.} We use classifier-free guidance with $w=4.0$, the lass label is ``Lakeside, lakeshore' (975).}
\end{figure*}

\end{document}